\documentclass{article}

\usepackage{microtype}
\usepackage{graphicx}
\usepackage{subfigure}
\usepackage{booktabs} %

\usepackage{hyperref}

\usepackage[accepted]{icml2025}

\usepackage{amsmath}
\usepackage{amssymb}
\usepackage{mathtools}
\usepackage{amsthm}

\usepackage{comment}
\usepackage{tikz}

\usepackage{pifont}
\usepackage{multirow}
\usepackage{bm}

\usepackage{xcolor}
\definecolor{pastelpink}{RGB}{255, 230, 235}

\newcommand{\highlightbf}[1]{\colorbox{pastelpink}{$\bm{#1}$}}
\newcommand{\highlightul}[1]{\colorbox{pastelpink}{\underline{#1}}}

\usepackage[capitalize,noabbrev]{cleveref}

\theoremstyle{plain}

\theoremstyle{definition}

\theoremstyle{remark}

\usepackage[textsize=tiny]{todonotes}

\hypersetup{colorlinks,linkcolor={blue},citecolor={magenta},urlcolor={blue}}

\icmltitlerunning{Mitigating Plasticity Loss in Continual RL by Reducing Churn}

\begin{document}

\twocolumn[
\icmltitle{Mitigating Plasticity Loss in Continual Reinforcement Learning \\ by Reducing Churn}

\begin{icmlauthorlist}
\icmlauthor{Hongyao Tang}{mila,udem}
\icmlauthor{Johan Obando-Ceron}{mila,udem}
\icmlauthor{Pablo Samuel Castro}{mila,udem}
\icmlauthor{Aaron Courville}{mila,udem}
\icmlauthor{Glen Berseth}{mila,udem}
\end{icmlauthorlist}

\icmlaffiliation{mila}{Mila - Qu\'{e}bec AI Institute}
\icmlaffiliation{udem}{Universit\'{e} de Montr\'{e}al}

\icmlcorrespondingauthor{Hongyao Tang}{tang.hongyao@mila.quebec}
\icmlcorrespondingauthor{Glen Berseth}{glen.berseth@mila.quebec}

\icmlkeywords{Machine Learning, ICML}

\vskip 0.3in
]

\printAffiliationsAndNotice{}  %

\begin{abstract}
Plasticity, or the ability of an agent to adapt to new tasks, environments, or distributions, is crucial for continual learning.
In this paper, we study the loss of plasticity in deep continual RL from the lens of churn: network output variability for out-of-batch data induced by mini-batch training.
We demonstrate that (1) the loss of plasticity is accompanied by the exacerbation of churn due to the gradual rank decrease of the Neural Tangent Kernel (NTK) matrix; (2) reducing churn helps prevent rank collapse %
and adjusts the step size of regular RL gradients adaptively.
Moreover, we introduce Continual Churn Approximated Reduction (C-CHAIN) and demonstrate it improves learning performance and outperforms baselines in a diverse range of continual learning environments on OpenAI Gym Control, ProcGen, DeepMind Control Suite, and MinAtar benchmarks.
\end{abstract}

\section{Introduction}
\label{sec:intro}

Reinforcement learning (RL), when coupled with non-linear function approximators, suffers from optimization challenges due to the non-stationarity of the data and the learning objectives~\citep{Sutton1988ReinforcementLA,Hasselt18DRLDeadly}. This difficulty is amplified when there is a sequence of changing tasks, as in continual RL~\citep{Abel0RPHS23ADefofCRL}.

One of the causes for this is what is known as \textit{loss of plasticity}~\citep{Berariu2021AStudyonPlasticity,LyleRD22Understanding,DohareHLRMS24LossofPlasticity}, whereby an agent gradually loses its ability to adapt to new data or objective function.
It is hypothesized that this is due to the agent's network tendency to overfit on early experience, hampering its ability to learn on later experience~\citep{NikishinSDBC22PrimacyBias}.
Addressing this pathology is challenging, and there have been a number of recently proposed remedies, such as resetting dormant neurons~\citep{Sokar2302Dormant}, regularization~\citep{Kumar2023MaintainingPI}, and variants of backpropagation~\citep{DohareHLRMS24LossofPlasticity}.
It has also been argued that rank decrease of the empirical Neural Tangent Kernel (NTK)~\citep{Achiam2019Towardscharacterizing} is a consistent indicator of plasticity loss~\citep{Lyle2024DisentanglingTC,Lewandowski2023DirectionsOC}. However, the true causal factors and mechanism underlying remain largely unknown.

In this work,
we study the loss of plasticity from the angle of \textit{churn}~\citep{Schaul22PolicyChurn,Tang2024ReduceChurn}: network output variability for out-of-batch data induced by mini-batch training.
While it has been shown that churn is endemic to most deep RL networks, affecting both generalization~\citep{BengioPP20Interference} and interference~\citep{LiuWTJ0W23Measuring},
we demonstrate that plasticity loss is highly correlated with increased churn.
We use the NTK matrix as a formal tool to establish the connection between churn and plasticity loss, based on which we analyze the learning dynamics of continual learning across a sequence of tasks.
We demonstrate that under the continual changes in the data distribution and objective function, the agent gradually loses the rank information of its NTK matrix, leading to highly correlated gradients and eventually the exacerbation of churn.
Consequently, this destroys the stability and convergence of the learning dynamics, which is related to the pathological learning behavior of plasticity loss.

Driven by the connection between plasticity and churn, we extend the idea of churn reduction originally proposed for single-MDP RL~\citep{Tang2024ReduceChurn} to continual RL.
We introduce Continual Churn Approximated Reduction (C-CHAIN).
C-CHAIN continually minimizes the churn for the data out of the training batch alongside the regular training of continual RL.
Moreover, we formally demonstrate that C-CHAIN has a two-fold efficacy in mitigating plasticity loss with a gradient decorrelation effect by suppressing the off-diagonal entries of the NTK matrix, and a step-size adjustment effect by taking the gradient projection of the data out of the training batch.

To evaluate our approach under extreme non-stationarity, we conduct the experiments in various continual RL environments built on OpenAI Gym Control~\citep{Brockman2016Gym}, ProcGen~\citep{CobbeHHS20ProcGen}, DeepMind Control Suite~\citep{Tassa2018DMC}, and MinAtar~\citep{Young2019MinAtar} benchmarks,
with a total of 24 continual RL environments. The results show that reducing churn effectively improves the agent's performance in continual RL, and outperforms related methods in most environments.

Our main contributions can be summarized as:
\begin{itemize}
    \item We demonstrate the connection between plasticity loss and increased churn, and show the pathological learning dynamics this connection induces.%
    \item We unbox the efficacy of reducing churn in continual RL by identifying a gradient decorrelation effect and a step-size adjustment effect.
    \item We propose C-CHAIN\footnote{\url{https://github.com/bluecontra/C-CHAIN}} and demonstrate it effectively mitigates the loss of plasticity and outperforms prior methods in a range of continual RL settings.
\end{itemize}

\section{Related Work}
\label{sec:related_work}

\paragraph{Nonstationarity in Reinforcement Learning}
Non-stationarity naturally occurs in various contexts, with continual RL being among the most prominent examples. Non-stationarity can manifest as gradual changes. This may happen when an RL agent continuously improves its policy over time \citep{zhang2024tackling} or when the data generation process slowly evolves \citep{Ellis2024AdamOnLocalTime}. These different forms of non-stationarity can coexist. For example, in continual reinforcement learning \citep{xie2021deep}, both the reward function and the transition dynamics may change over time, requiring learning systems to adapt dynamically.

Due to the use of a non-linear function approximator under non-stationarity, the model weights and optimization may not lead to an optimal policy. There are two main challenges in optimizing under distribution shift, catastrophic forgetting (also called \textit{backward transfer})~\citep{RolnickASLW19ERCL,ChaudhryRRE19AGEM,NathPBLDLKS23SharingLifelongKnowledge} or 
the loss of plasticity (called \textit{forward transfer})~\citep{Berariu2021AStudyonPlasticity,LyleRD22Understanding,DohareHLRMS24LossofPlasticity}. 

\vspace{-0.3cm}
\paragraph{Continual Learning and the Loss of Plasticity}
Continual learning focuses on training models to learn sequential tasks.
It is broadly divided into continual supervised learning, which aims to retain knowledge across tasks in static datasets \cite{wang2024comprehensive}, and continual RL, where agents adapt to dynamic environments while maintaining prior experience \citep{khetarpal2022towards,kumar2023continual,DohareHLRMS24LossofPlasticity,elsayed2024addressing}.
The loss of plasticity refers to a phenomenon in neural network training where the model gradually loses its ability to adapt or significantly update its parameters \cite{DohareHLRMS24LossofPlasticity}. This typically occurs when the network settles into sharp local minima or becomes overly specialized to the initial stages of training, resulting in reduced learning capacity, slower convergence, or poor generalization to new data \cite{nikishin2022primacy,nikishin2024deep}. 

The underlying mechanisms responsible for the loss of plasticity during training remain an area of active research \citep{ma2023revisiting,Lyle2024DisentanglingTC}. Researchers have proposed various strategies to mitigate the loss of plasticity; approaches include ensuring active unit engagement \citep{Sokar2302Dormant}, mitigating gradient starvation \citep{gogianu2021spectral,Dohare23Maintaining}, employing smaller batch sizes \citep{ceron2023small}, minimizing deviations from initial parameter values \citep{Lewandowski2023DirectionsOC,kumar2023continual},
resetting optimizer~\citep{AsadiFS23ResetOpt,Ellis2024AdamOnLocalTime},
and regularizing weight orthogonality~\citep{chung2024parseval}. 
Periodically resetting network parameters has been effective \citep{ash2020warm,frati2024reset}, especially effective in data-efficient reinforcement learning \citep{schwarzer2023bigger,naumanoverestimation}.\citet{ceronvalue,liu2025neuroplastic} showed that dynamic sparse training leads to improved performance and enhanced network plasticity.

\section{Preliminaries}
\label{sec:preliminaries}

\subsection{Reinforcement Learning}
Consider a Markov Decision Process (MDP) defined by a tuple $\left< \mathcal{S}, \mathcal{A}, \mathcal{P}, \mathcal{R},  \gamma, \rho_0, T \right>$, with the state set $\mathcal{S}$, the action set $\mathcal{A}$, the transition function $\mathcal{P}: \mathcal{S} \times \mathcal{A} \rightarrow P(\mathcal{S})$, the reward function $\mathcal{R}: \mathcal{S} \times \mathcal{A} \rightarrow \mathbb{R}$, the discounted factor $\gamma \in [0,1)$, the initial state distribution $\rho_0$ and the horizon $T$.
The agent interacts with the MDP by performing actions with its policy $a_t \sim \pi(s_t)$ that defines the mapping from states to action distributions.
The objective of an RL agent is to optimize its policy to maximize the expected discounted cumulative reward
$J(\pi) = \mathbb{E}_{\pi} [\sum_{t=0}^{T}\gamma^{t} r_t ]$,
where $s_{0} \sim \rho_{0}\left(s_{0}\right)$, $s_{t+1} \sim \mathcal{P}\left(s_{t+1} \mid s_{t}, a_{t}\right)$ and $r_t = \mathcal{R}\left(s_{t},a_{t}\right)$.
The state-action value function $q^{\pi}$ defines the expected cumulative discounted reward for all $s,a \in \mathcal{S} \times \mathcal{A}$ and the policy $\pi$,
i.e., $q^{\pi}(s, a)=\mathbb{E}_{\pi} \big[\sum_{t=0}^{T} \gamma^{t} r_{t} \mid s_{0}=s, a_{0}=a \big]$.

In deep RL, policy and value functions are approximated with deep neural networks,
conventionally denoted by $Q_{\theta}$ and $\pi_{\phi}$ with network parameters $\theta$ and $\phi$.
The parameterized policy $\pi_{\phi}$ can be updated by taking the gradient of the objective, 
i.e., $\phi^{\prime} \leftarrow \phi + \alpha \nabla_{\phi} J(\pi_{\phi})$ with a step size $\alpha$~\citep{Silver2014DPG,MnihBMGLHSK16A3C,SchulmanWDRK17PPO,HaarnojaZAL18SAC}.

\subsection{Continual Learning and Plasticity}

Beyond the standard MDP setting which describes a stationary decision-making task, we consider a continual learning scenario where the agent learns to solve tasks $\mathbb{T}=\{\mathcal{T}_1, \mathcal{T}_2, ..., \mathcal{T}_k\}$ that arrive in a sequence.
In this work, each task is an MDP with a different reward or dynamics function and has a budget of $N$ interactions for the agent.
Aside from this, we do not use any assumption on the task sequence, e.g., the similarity between tasks.
The non-stationarity manifests in the continual switch of learning tasks. 
We focus on plasticity in this work, and the objective of a continual RL agent in this context is to maximize the \textit{average performance} over the course of continual learning on the task sequence.

Formally, for any point $i$ within the total budget, we have the policy $\pi_{i}$ where $i \in \{1, ..., k\}$, and the performance $J(\pi_{i})$ corresponding to it.
Note that $J(\pi_{i})$ is the performance evaluated on the task $\mathcal{T}_{\lceil i \rceil}$, i.e., the current task at time step $i$.
For convenience, we define the average performance regarding a task $\mathcal{T}_j$ as $J(\mathcal{T}_j) = \frac{1}{N} \sum_{i=(j - 1)*N + 1}^{j*N} J(\pi_{i})$.
Thus, the average performance of the agent throughout the learning on the sequence $\mathbb{T}$ is:
\begin{equation}
    J(\mathbb{T}) = \frac{1}{k} \sum_{j=1}^{k} J(\mathcal{T}_j) = \frac{1}{kN} \sum_{i=1}^{kN} J(\pi_i).
\label{eq:crl_avg_performance_objective}
\end{equation}
Similar to the concept of Area Under Curve (AUC), the average performance metric aggregates the learning performance in terms of efficiency, stability and convergence.
We use average performance to evaluate and compare different methods, and in practice, we take intervals within $\{1, ..., k\}$ for the estimation of $J(\mathbb{T})$.

In the context of continual learning, the loss of plasticity means that the agent gradually exhibits worse learning performance on later tasks.
Empirically, it can often be identified when we observe $\bar J(\mathcal{T}_i) - J(\mathcal{T}_i) \ge \bar J(\mathcal{T}_j) - J(\mathcal{T}_j)$ for $1 \le i \le j \le k$, where $\bar J(\mathcal{T}_i)$ denotes the performance of an agent of good plasticity on the task $\mathcal{T}_i$, e.g., a newly initialized network for the current task.

\section{Mitigating the Loss of Plasticity by Reducing Churn}
\label{sec:method}

In this section, we formally study the loss of plasticity in continual RL from the lens of churn.
First, we establish a connection between plasticity loss and churn by taking the NTK matrix as a framework (Section ~\ref{subsec:bridge_between_plas_and_churn}), 
based on which we analyze the continual learning dynamics (Section~\ref{subsec:learning_dynamics}).
Further, we 
introduce Continual Churn Approximated Reduction (C-CHAIN), and formally dissect the efficacy of churn reduction on plasticity
(Section~\ref{subsec:dissecting_the_effect_of_churn_reduction}).

\subsection{NTK as the Bridge between Plasticity and Churn}
\label{subsec:bridge_between_plas_and_churn}

The loss of plasticity is a phenomenon that stems from the pathological learning behavior of a deep network, while churn is an innate feature of the network where the network output for datapoints not included in the current training batch is implicitly and potentially uncontrollably changed.
To study the relationship between the two concepts, we use the empirical NTK~\citep{Achiam2019Towardscharacterizing} matrix as a formal tool, as its rank has been shown to be a good indicator of plasticity loss~\citep{Lyle2024DisentanglingTC}.

For a parameterized network function $f_{\theta}(x) \in \mathbb{R}$, we use $g(x)$ to denote the gradient of $f_{\theta}$ with respect to its parameters $\theta$ at the data $x$, i.e., $g(x) = \nabla_{\theta}f_{\theta}(x) \in \mathbb{R}^{d}$ where $|\theta|=d$ is the dimensionality.
The Neural Tangent Kernel matrix $N_{\theta}$ is the matrix of gradient dot products between all data points~\citep{Achiam2019Towardscharacterizing,Lyle2024DisentanglingTC}:
\begin{equation}
\begin{aligned}
    N_{\theta}(i,j) & = \nabla_\theta f_{\theta}(x_i)^\top \nabla_{\theta} f_{\theta}(x_j) \ \ \text{for} \ x_i, x_j \\
    N_{\theta} & = G_{\theta}^{\top}G_{\theta}
\end{aligned}
\label{eq:ntk_def}
\vspace{-0.1cm}
\end{equation}
where $G_{\theta} = [g(x_1), g(x_2), \dots, g(x_i), \dots]$ denotes the matrix form of the gradients for all datapoints.
This is also referred to as the outer-product approximation of Hessian matrix~\citep{Lewandowski2023DirectionsOC}.
In practice, the empirical NTK matrix is often used by sampling a batch of datapoints for the convenience of computation and visualization. \citet{Lyle2024DisentanglingTC} show that networks tend to exhibit a low rank of the empirical NTK matrix {$N_{\theta}$} when they lose plasticity.

\begin{figure}
\centering
\includegraphics[width=0.6\linewidth]{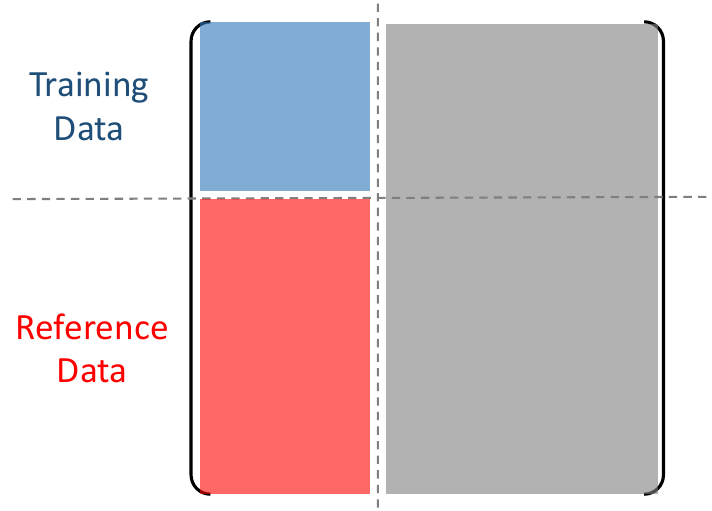}
\vspace{-0.4cm}
\caption{\textbf{An illustration of the matrix $N_{\theta} S$}, where the datapoints are arranged with a separation between the training data and the remaining ones (i.e., the reference data).
}
\label{figure:ntk_illustration}
\vspace{-0.4cm}
\end{figure}

Now consider the parameter changes from $f_{\theta}$ to $f_{\theta^{\prime}}$ and let $\Delta_{\theta} = \theta^{\prime} - \theta$.
For data $\bar x \in B_{\text{ref}}$ (the \textit{reference} data) not included in training batch $B_{\text{train}}$ used for the parameter update $\Delta_{\theta}$, the churn of $f_{\theta}$ at $\bar x$ with respect to $\Delta_{\theta}$ is~\citep{Tang2024ReduceChurn}:
\begin{equation}
\begin{aligned}
    C_f(\bar x, \theta, \Delta_{\theta}) & = f_{\theta^{\prime}}(\bar x) - f_{\theta}(\bar x) \\
    & = \nabla_{\theta}f_{\theta}(\bar x)^{\top} \Delta_{\theta} + O(\|\Delta_{\theta}\|^{2}).
\end{aligned}
\label{eq:churn_def}
\vspace{-0.1cm}
\end{equation}
The parameter update $\Delta_{\theta}$ made by mini-batch training can have different forms, which depends on the context.
Letting $L(\theta)$ be a loss function of $f_{\theta}$, we can express the parameter update in a general form via the chain rule: $\Delta_{\theta} = - \eta \mathbb{E}_{x} [\nabla_{\theta}f_{\theta}(x) \nabla_{f_{\theta}} L({\theta, x})]$, where $x \in B_{\text{train}}$ is the data sampled for the mini-batch training. We use vanilla stochastic gradient descent and use $\eta$ for the learning rate here for legibility.
By plugging the general form back, we have the NTK expression of the churn:
\begin{equation}
\begin{aligned}
    C_f(\bar x, \theta, \Delta_{\theta}) & \approx - \eta \nabla_{\theta}f_{\theta}(\bar x)^{\top} \mathbb{E}_{x} [\nabla_{\theta}f_{\theta}(x) \nabla_{f_{\theta}} L({\theta, x})] \\
    & = - \eta \mathbb{E}_{x} [ {N_{\theta}(\bar x, x)} \nabla_{f_{\theta}} L({\theta, x})].
\end{aligned}
\label{eq:churn_ntk}
\end{equation}
Further, we can obtain the churn in vector form, which is equivalent to the approximate change of the function value for all datapoints:
\begin{equation}
\begin{aligned}
    C_f(\theta, \Delta_{\theta}) & \approx - \eta G_{\theta}^{\top} G_{\theta}  S G_L  \\
    & = - \eta {N_{\theta}}  S G_L ,
\end{aligned}
\label{eq:churn_vector_form}
\vspace{-0.1cm}
\end{equation}
where $G_L$ is the gradient matrix of $\nabla_{f_{\theta}}L(\theta)$ and $S$ is a diagonal matrix of the same size as $N_{\theta}$ with $\{0,1\}$-binary values on its diagonal that corresponds to the sampling results of $B_{\text{train}}$.
The vector form in Equation~\ref{eq:churn_vector_form} indicates that the NTK matrix $N_{\theta}$ plays an important role in determining the churn, independent of the loss function (or objective function) and the sampling strategy (or data distribution).
Intuitively, $N_{\theta}$ determines how the explicit mini-batch training shapes the entire function landscape by generalization or interference between different datapoints.
Figure~\ref{figure:ntk_illustration} shows an illustration of the matrix $N_{\theta} S$. For the symmetric NTK matrix $N_{\theta}$, multiplying the sampling matrix $S_i$ zero-masks out the columns corresponding to the reference data, denoted by the \scalebox{0.5}[0.5]{\colorbox{gray}{\phantom{A}}} area. The remaining entries consist of the training updates to the sampled training data (\scalebox{0.5}[0.5]{\colorbox{blue}{\phantom{A}}}) and the changes caused by churn to the reference data (\scalebox{0.5}[0.5]{\colorbox{red}{\phantom{A}}}).

As the empirical NTK matrix plays important roles in both plasticity loss and churn, it serves as a natural bridge between the two.
We further our analysis below.

\subsection{Exacerbation of Churn Induced by NTK Collapse}
\label{subsec:learning_dynamics}

Starting from a random parameter initialization, the network of a learning agent digests data from the sequential tasks $\mathbb{T}=\{\mathcal{T}_1, \mathcal{T}_2, ..., \mathcal{T}_k\}$ and updates it parameters to solve each task at hand the best.
Notice that each task is an MDP and the data on the task is collected online by the agent within a budget of interaction steps.

To formally characterize the learning process, let $\mathcal{E}_{i}(\theta)=f^{\ast}_{i} - f_{\theta}$ be the error vector of the network $f_{\theta}$ regarding the optimal objective function $f^{\ast}_{i}$ corresponding to task $\mathcal{T}_i$ with $i \in \{1, \dots, k\}$.
We consider the commonly-used loss function $L_{i}({\theta, x}) = \frac{1}{2} [f^{\ast}_{i}(x) - f_{\theta}(x)]^2$ .
The dynamics of the error $\mathcal{E}_{i}$ for iterative updates $\theta_{t} \rightarrow \theta_{t+1}$ within the same task $\mathcal{T}_i$ can be derived as:
\begin{equation}
\begin{aligned}
    \mathcal{E}_{i}(\theta_{t+1}) - \mathcal{E}_{i}(\theta_t) & = \big( f^{\ast}_{i} - f_{\theta_{t+1}} \big) - \big( f^{\ast}_{i} - f_{\theta_{t}} \big) \\
    & = f_{\theta_{t}} - f_{\theta_{t+1}} \\
    & = - C_f(\theta_t, \theta_{t+1} - \theta_{t}) .
\end{aligned}
\label{eq:error_diff}
\vspace{-0.1cm}
\end{equation}
By plugging the vector form of churn in Equation~\ref{eq:churn_vector_form} and $G_{L} = - \big( f^{\ast}_{i} - f_{\theta_t} \big) = - \mathcal{E}_i(\theta_t)$, we further have:
\begin{equation}
\begin{aligned}
    \mathcal{E}_{i}(\theta_{t+1}) & \approx \mathcal{E}_{i}(\theta_t) - \eta {N_{\theta_t}}  S_i \mathcal{E}_{i}(\theta_t)   \\
    & = \big(I - \eta {N_{\theta_t}}  S_i \big) \mathcal{E}_{i}(\theta_t) .
\end{aligned}
\label{eq:error_dynamaics}
\vspace{-0.1cm}
\end{equation}
Equation~\ref{eq:error_dynamaics} characterizes the iterative evolution of the error regarding the task $\mathcal{T}_i$, which is mainly determined by the factor $I - \eta {N_{\theta_t}}  S_i $.
$N_{\theta_t}$ is full-rank when it has positive diagonal values (i.e., non-zero gradients $f_{\theta}(x_i) \ne \mathbf{0}$) and has zero off-diagonal entries, i.e., $\nabla_\theta f_{\theta}(x_i)^\top \nabla_{\theta} f_{\theta}(x_j) = 0$
for $i \ne j$.
In this case, it resembles the tabular function approximation where generalization and interference among different datapoints do not exist.
The learning process is stable when a proper $\eta$ is selected.
However, in most practical cases, the off-diagonal entries of $N_{\theta_t}$ are non-zero and $N_{\theta_t}$ is not full-rank, making the learning dynamics intricate.

Next, we continue to analyze the learning dynamics by taking into consideration the task change in $\mathbb{T}=\{\mathcal{T}_1, \mathcal{T}_2, ..., \mathcal{T}_k\}$.
Different from a stationary learning scenario, both the data distribution (related to $S_i$) and the objective function (related to $f_i^{\ast}, \mathcal{E}_i$) change throughout learning.
In the early stages of learning, the parameters of the agent's network are close to the initialization, thus having no plasticity issue. Meanwhile, the empirical objective function estimated with finite online data tends to lie in low-dimensional space.
With stochastic gradient descent, the network fits the empirical objective function with an implicit preference for simple functions, i.e., the Simplicity Bias~\citep{ArpitJBKBKMFCBL17ACloseLook,KalimerisKNEYBZ19SGDonNN,DamianML21LabelNoise}.
In this process, the agent prefers the parameters that have high correlations between different datapoints~\citep{KumarA0CTL22DR3} in order to generalize the updates, and loses the rank information in $N_{\theta}$ since the empirical objective function (related to $S_i, \mathcal{E}_i$) for the current task is low-dimensional.

When the task changes from $\mathcal{T}_i$ to $\mathcal{T}_{i+1}$, the agent continues to fit the new objective function from $\mathcal{E}_{i+1}(\theta)$ with a distribution shift of sampling data $S_{i+1}$.
However, the function landscape regarding the new data distribution was implicitly shaped via churn in prior training. Compared to the parameters in the early stage, the gradients between different datapoints correlate more and the rank of $N_{\theta}S_{i+1}$ becomes lower.
This echoes the findings in prior works~\citep{Lyle2024DisentanglingTC,KumarAGL21ImplicitUnder,ShahTR0N20PitfallofSB}.
As the task changes along the sequence $\mathbb{T}$, the rank decrease of the NTK matrix and the exacerbation of churn operate in a vicious cycle in the learning dynamics (Equation~\ref{eq:error_dynamaics}).
Therefore, it leads to even less stable learning and worse approximation results, which is the pathological learning behavior identified as the loss of plasticity.

Naturally, a method that prevents the rank decrease of $N_{\theta}S_i$ and the exacerbation of churn should mitigate the loss of plasticity.
Therefore, we dissect the efficacy of reducing churn on continual learning in the next subsection.

\subsection{Effects of Continual Churn Reduction}
\label{subsec:dissecting_the_effect_of_churn_reduction}

We extend the study on churn reduction and explore its potential in mitigating plasticity loss in continual RL.
We introduce Continual Churn Approximated Reduction (C-CHAIN).
As depicted by the pseudo-code in Algorithm~\ref{alg:c_chain}, the agent interacts with the tasks that arrive in a sequence.
Alongside the regular training of the continual RL algorithm, C-CHAIN continually minimizes the churn for the data out of the current training batch (i.e., the reference batch $B_{\text{ref}}$).

Formally, the churn reduction loss function of C-CHAIN is defined with the reference data $\bar x \in B_{\text{ref}}$ as follows:
\vspace{-0.1cm}
\begin{equation}
\begin{aligned}
    L_{f}^{\text{cr}}(\theta) & = \frac{1}{2}\mathbb{E}_{\bar x \in B_{\text{ref}}} [C_{f} (\bar x,\theta,\Delta_{\theta})^2]
\end{aligned}
\label{eq:churn_reduction_loss}
\vspace{-0.1cm}
\end{equation}
Next, we look into the gradient of the churn reduction loss to shed light on how it influences the continual learning dynamics we discussed in the previous subsection:
\begin{equation}
\begin{aligned}
    \nabla_{\theta}L_{f}^{\text{cr}}(\theta, \bar x) & = C_f(\bar x, \theta, \Delta_{\theta})\nabla_{\theta} C_f(\bar x, \theta, \Delta_{\theta}) \\
    \nabla_{\theta} C_f(\bar x, \theta, \Delta_{\theta}) & \approx  - \eta \nabla_{\theta} \big( \mathbb{E}_{x} [ {N_{\theta}(\bar x, x)} \nabla_{f_{\theta}} L({\theta, x})] \big)
\end{aligned}
\label{eq:grad_of_churn}
\end{equation}
For clarity we use $g = \nabla_{\theta}f_{\theta}(x), \bar g = \nabla_{\theta}f_{\theta}(\bar x)$, thus the kernel $N_{\theta}(\bar x, x) = \bar g^{\top} g$, and $\nabla_f L_f = \nabla_{f_{\theta}} L({\theta, x})$.
We then have the derivative in the expectation below:
\begin{equation}
\begin{aligned}
    \nabla_{\theta} \big( {\bar g^{\top} g} \nabla_f L_f \big) = \underbrace{\nabla_{\theta} \big( {\bar g^{\top} g}  \big) \nabla_f L_f}_{\text{\ding{172}}} + \underbrace{ {\bar g^{\top} g} \nabla_{\theta} \big( \nabla_f L_f \big) }_{\text{\ding{173}}}
\end{aligned}
\label{eq:separate_derivative}
\end{equation}
The \ding{172} term is the partial derivative of the gradients of $f_{\theta}$. With $H_{\theta}$ be the Hessian matrix of $f_{\theta}$, we have:
\begin{equation}
\begin{aligned}
    \text{\ding{172}} = \big( H_{\theta}(\bar x) g + H_{\theta}(x) \bar g \big) \nabla_f L_f
\end{aligned}
\label{eq:the_first_term}
\vspace{-0.1cm}
\end{equation}
Equation~\ref{eq:the_first_term} indicates the first efficacy of reducing churn.
It takes the gradient of the kernel $N_{\theta}(\bar x, x)$ regarding each pair of the reference data $\bar x$ and the training data $x$, with $\nabla_f L_{f}$ (as well as $C_f(\bar x, \theta, \Delta_{\theta}$) playing the role of a weighting factor.
As a result, the off-diagonal entries in the NTK matrix $N_{\theta}$ (corresponding to the \scalebox{0.5}[0.5]{\colorbox{red}{\phantom{A}}} area of Figure~\ref{figure:ntk_illustration}) are suppressed to zero for minimization.

\begin{algorithm}[t]
    \small
    \begin{algorithmic}[1]
        \STATE Task sequence $\mathbb{T}=\{\mathcal{T}_1, \mathcal{T}_2, ..., \mathcal{T}_k\}$, interaction budget per task $N$, continual RL algorithm $\mathbb{A}$
        \STATE Initialize agent's parameter $\theta$ and a buffer $D$
        \FOR{task $i  =  1, 2, 3, ..., k$}
            \FOR{step $i  =  1, 2, 3, ..., N$}
                \STATE Interact with the task by performing agent's policy
                \STATE Update the buffer $D$ according to $\mathbb{A}$
                \STATE Perform regular parameter update $\Delta_{\theta}=\mathbb{A}(\theta, B_{\text{train}})$ with the training batch $B_{\text{train}} \sim D$
                \STATE Regularize agent's parameter to reduce churn for the reference batch $B_{\text{ref}} \sim D$ ($B_{\text{ref}} \cap B_{\text{train}} = \emptyset$)
            \ENDFOR
        \ENDFOR
        \end{algorithmic}
    \caption{
    Deep Continual RL with Continual Churn Approximated Reduction (C-CHAIN).
    }
\label{alg:c_chain}
\end{algorithm}

The \ding{173} term depends on the specific form of $L_{f}$ regarding which the regular training (i.e., $\Delta_{\theta}$) is performed.
Here we use the Temporal Difference (TD) learning of $Q$-network for $L_{f}$ as a common example in deep RL for demonstration, where $L_{Q}(\theta) = \frac{1}{2} \mathbb{E}_{x,x^{\prime}} [\big( (r+\gamma Q^{-}(x^{\prime})) - Q_{\theta}(x) \big)^2]$.
Note that we use $x$ for $s,a$ and $Q^{-}$ is the target $Q$-function detached from backpropagation.
In this case, we have $\nabla_{Q_{\theta}} L_{Q}(\theta, x) = -Q_{\theta}(x) + (r+\gamma Q^{-}(x^{\prime}))$ and thus:
\begin{equation}
\begin{aligned}
    \text{\ding{173}} = - (\bar g^{\top} g) g = - \textbf{proj}_{g} \bar g \ \|g\|_2^{2}
\end{aligned}
\label{eq:the_second_term}
\end{equation}
Equation~\ref{eq:the_second_term} shows that the second efficacy of reducing churn can be interpreted as the projection of $\bar g$ on $g$ with the square norm of $g$ as a scaling factor.
This term could either dampen or accelerate the regular gradient $g \nabla_f L_f$ which depends on the sign of the kernel between the reference data and the training data, and also the $\nabla_f L_f$.

The two terms work together when reducing churn during the continual learning process.
The \ding{172} term decorrelates the gradients between the reference data and the training data by suppressing the off-diagonal entries in the NTK matrix $N_{\theta}$.
This combats the rank decrease and mitigates the loss of plasticity under the changes of the data distribution and the objective function as discussed in Section~\ref{subsec:learning_dynamics}.
Besides, the \ding{173} term regulates the regular training gradient based on additional gradient information of the reference data.
More regularization occurs when the gradients correlates more.

To summarize, reducing churn improves the stability of the continual learning dynamics in Equation~\ref{eq:error_dynamaics} and mitigates the loss of plasticity.
In the next section, we conduct the empirical study for our formal findings discussed above.

\section{Experiments}
\label{sec:experiments}

\begin{table*}
  \vspace{-0.3cm}
  \caption{\textbf{Performance comparison on continual Gym Control.} 
  The reported scores are the average performance (Eq.~\ref{eq:crl_avg_performance_objective}) in terms of episode return with mean and standard error over six seeds. 
  The top-$2$ values (excluding Oracle) are marked by \textbf{bold} and \underline{underline}.}
  \centering
  \scalebox{0.7}{
  \begin{tabular}{c|c|c|c|c|c|c}
    \toprule
    \multirow{2}{*}{Env\vspace{-4pt}} & \multicolumn{2}{c|}{Calibration Baselines} & \multicolumn{3}{c|}{Related Methods} & Ours \\
    \cmidrule{2-7}
    & Oracle & Vanilla & TRAC & Weight Clipping & L2 Init & C-CHAIN \\
    \midrule
    C-Acrobot & -125.736 $\pm$ 8.303 & -372.725 $\pm$ 3.268 & -166.584 $\pm$ 8.922 & \highlightbf{-118.821 \pm 13.440} & -134.038 $\pm$ 14.292 & \highlightul{-119.211 $\pm$ 6.929}  \\
    C-CartPole & 299.925 $\pm$ 5.986 & 61.766 $\pm$ 6.927 & 217.376 $\pm$ 26.307 & 154.306 $\pm$ 16.047 & \highlightbf{266.114 \pm 5.345} & \highlightul{160.396 $\pm$ 14.643} \\
    C-LunarLander & 1.118 $\pm$ 2.002 & -499.396 $\pm$ 19.571 & -58.059 $\pm$ 17.474 & -58.575 $\pm$ 2.896 & \highlightbf{-6.666 \pm 5.903} & \highlightul{-16.659 $\pm$ 8.369} \\
    C-MountainCar & -322.608 $\pm$ 9.384 & -346.604 $\pm$ 4.635 & -399.989 $\pm$ 0.010 & \highlightul{-267.900 $\pm$ 6.883} & -369.395 $\pm$ 27.914 & \highlightbf{-245.746 \pm 14.691} \\
    \midrule
    Agg. Score & -147.302 & -1156.958 & -407.256 & -290.99 & \highlightul{-243.985} & \highlightbf{-221.22} \\
    \bottomrule
  \end{tabular}
  }
\label{table:crl_control_results_summary}
\vspace{-0.2cm}
\end{table*}

\begin{table*}
  \vspace{-0.3cm}
  \caption{\textbf{Performance comparison on continual ProcGen.} 
  The reported scores are the average performance (Eq.~\ref{eq:crl_avg_performance_objective}) in terms of episode return with mean and standard error over six seeds. 
  The top-$2$ values (excluding Oracle) on each row are marked by \textbf{bold} and \underline{underline}.}
  \centering
  \scalebox{0.66}{
  \begin{tabular}{c|c|c|c|c|c|c|c|c|c}
    \toprule
    \multirow{2}{*}{Env\vspace{-4pt}} & \multicolumn{2}{c|}{Calibration Baselines} & \multicolumn{6}{c|}{Related Methods} & Ours \\
    \cmidrule{2-10}
    & Oracle & Vanilla & TRAC & Weight Clip & L2 Init & LayerNorm & ReDo & AdamRel & C-CHAIN \\
    \midrule
    Starpilot & 13.142 $\pm$ 0.262 & 8.028 $\pm$ 0.667 & \highlightul{18.106 $\pm$ 0.716} & 8.200 $\pm$ 0.729 & 11.353 $\pm$ 0.207 &  14.085$\pm$0.652 & 9.824$\pm$0.221 & 11.863$\pm$0.910 & \highlightbf{19.717 \pm 0.699}  \\
    Fruitbot & 1.613 $\pm$ 0.105 & 1.549 $\pm$ 0.139 & 0.855 $\pm$ 0.327 & 1.457 $\pm$ 0.090 & \highlightul{1.657 $\pm$ 0.107} & 0.921$\pm$0.254 & 1.586$\pm$0.227 & 1.487$\pm$0.149 &  \highlightbf{1.689 \pm 0.188}  \\
    Chaser & 2.923 $\pm$ 0.016 & 2.714 $\pm$ 0.028 & \highlightul{3.096 $\pm$ 0.021} & 2.812 $\pm$ 0.039 & 3.067 $\pm$ 0.023 & 2.960$\pm$0.034 & 2.894$\pm$0.032 & 2.720$\pm$0.053 &  \highlightbf{3.432 \pm 0.033} \\
    Dodgeball & 5.017 $\pm$ 0.415 & 3.636 $\pm$ 0.488 & \highlightul{6.757 $\pm$ 0.617} & 3.066 $\pm$ 0.600 & 4.859 $\pm$ 0.459 & 5.895$\pm$0.787 & 3.799$\pm$0.324 & 3.780$\pm$0.643 &  \highlightbf{7.690 \pm 0.840}  \\
    Bigfish & 4.206 $\pm$ 0.343 & 2.363 $\pm$ 0.632 & \highlightul{5.783 $\pm$ 0.729} & 2.561 $\pm$ 0.334  & 3.752 $\pm$ 0.267 & 4.406$\pm$0.434 & 3.010$\pm$0.199 & 2.746$\pm$0.343  & \highlightbf{7.866 \pm 0.319} \\
    Caveflyer & 5.889 $\pm$ 0.285 & 5.605 $\pm$ 0.282 & \highlightul{6.760 $\pm$ 0.400} & 5.157 $\pm$ 0.362 & 5.922 $\pm$ 0.307 & 6.103$\pm$0.228 & 5.595$\pm$0.330 & 5.232$\pm$0.491 & \highlightbf{7.653 \pm 0.587}  \\
    Climber & 5.404 $\pm$ 0.659 & 4.557 $\pm$ 0.692 & 2.557 $\pm$ 0.500 & 5.003 $\pm$ 0.740  & \highlightul{5.415 $\pm$ 0.662} & 5.349$\pm$0.658 & 5.372$\pm$0.634 & 5.207$\pm$0.677 & \highlightbf{5.594 \pm 0.579} \\
    Ninja & 2.169 $\pm$ 0.311 & 1.679 $\pm$ 0.705 & 1.278 $\pm$ 0.556 & 1.314 $\pm$ 0.601 & \highlightbf{3.219 \pm 0.364} &  2.760$\pm$0.572 & 2.184$\pm$0.434 & 1.297$\pm$0.597 & \highlightul{3.146 $\pm$ 0.436} \\
    Coinrun & 4.953 $\pm$ 0.458 & 4.711 $\pm$ 0.676 & 3.943 $\pm$ 1.414  & 4.985 $\pm$ 0.757  & 4.857 $\pm$ 0.362 &  \highlightul{5.459$\pm$1.063} & 4.689$\pm$0.376 & 4.377$\pm$0.503 & \highlightbf{6.829 \pm 1.093} \\
    Miner & 5.147 $\pm$ 0.235 & 4.250 $\pm$ 0.314 & 4.288 $\pm$ 0.807 & 5.182 $\pm$ 0.634  & \highlightul{6.153 $\pm$ 0.228} &  5.491$\pm$0.542 & 4.853$\pm$0.397 & 5.500$\pm$0.645  & \highlightbf{9.924 \pm 0.636} \\
    Jumper & 2.175 $\pm$ 0.330 & 1.095 $\pm$ 0.211 & 1.753 $\pm$ 0.456 & 1.536 $\pm$ 0.474  & 2.598 $\pm$ 0.299 &  \highlightul{2.666$\pm$0.471} & 1.269$\pm$0.274 & 2.225$\pm$0.509  & \highlightbf{2.896 \pm 0.371} \\
    Heist & 2.426 $\pm$ 0.295 & 2.404 $\pm$ 0.290 & 2.424 $\pm$ 0.294 & 2.506 $\pm$ 0.295  & \highlightul{2.929 $\pm$ 0.343} &  2.402$\pm$0.292 & 2.691$\pm$0.368 & 2.668$\pm$0.372  & \highlightbf{3.950 \pm 0.681} \\
    Leaper & 0.350 $\pm$ 0.302 & 0.347 $\pm$ 0.302 & 0.334 $\pm$ 0.304 & 0.474 $\pm$ 0.305 & 0.557 $\pm$ 0.287 &  \highlightbf{0.809 \pm 0.443} & 0.467$\pm$0.303 & 0.511$\pm$0.314  & \highlightul{0.668 $\pm$ 0.608} \\
    Maze & 5.528 $\pm$ 0.525 & 5.324 $\pm$ 0.621 & 4.916 $\pm$ 0.774 & 5.314 $\pm$ 0.885 & \highlightul{5.701 $\pm$ 0.999} &  5.084$\pm$0.550 & 4.747$\pm$0.616 & 5.299$\pm$0.867  & \highlightbf{5.940 \pm 0.424} \\
    Plunder & 7.113 $\pm$ 0.413 & 4.801 $\pm$ 0.430 & \highlightul{9.719 $\pm$ 0.776} & 5.290 $\pm$ 0.389 & 6.695 $\pm$ 0.383 &  7.262$\pm$0.363 & 5.910$\pm$0.331 & 5.296$\pm$0.298  & \highlightbf{10.340 \pm 0.287} \\ 
    Bossfight & 2.735 $\pm$ 0.369 & 1.986 $\pm$ 0.667 & \highlightul{4.722 $\pm$ 1.027} & 2.235 $\pm$ 0.630 & 3.791 $\pm$ 0.436 &  3.503$\pm$0.320 & 2.290$\pm$0.352 & 1.541$\pm$0.506  & \highlightbf{4.920 \pm 0.45}\\
    \midrule
    Agg. Score & 70.789 & 55.049 & \highlightul{77.289} & 57.092 & 72.961 & 75.154 & 61.180 & 61.443   & \highlightbf{101.792}\\
    \bottomrule
  \end{tabular}
  }
\label{table:crl_procgen_results_summary}
\vspace{-0.5cm}
\end{table*}

In our experiments, we first evaluate C-CHAIN in comparison with recent related methods, to find out whether it improves continual RL (Section~\ref{subsec:performance_comparison}).
Then, we conduct the empirical analysis to examine whether reducing churn prevents the rank decrease and how the two effects contribute differently (Section~\ref{subsec:empirical_analysis}).
Finally, we extend the evaluation to more continual learning settings (Section~\ref{subsec:more_results}).

\subsection{Performance Evaluation}
\label{subsec:performance_comparison}

To evaluate the performance of continual RL, we adopt OpenAI Gym Control~\citep{Brockman2016Gym} and ProcGen~\citep{CobbeHHS20ProcGen} and follow the setups in TRAC~\citep{Muppidi2024TRAC}.

\begin{figure}
\begin{center}
\includegraphics[width=0.44\textwidth]{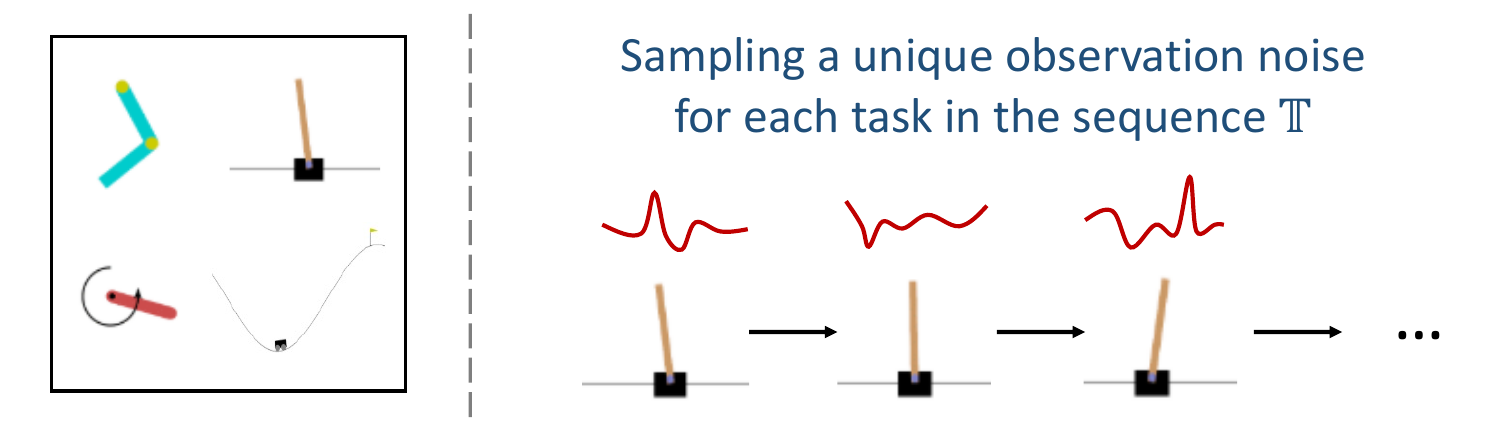}
\vspace{0.3cm}
\includegraphics[width=0.48\textwidth]{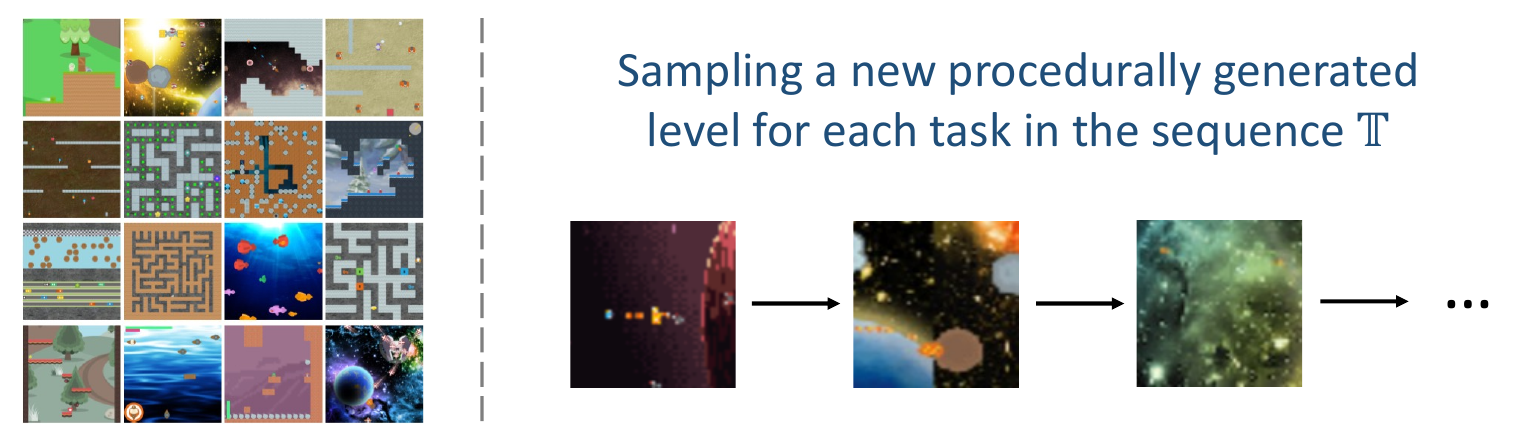}
\end{center}
\vspace{-1cm}
\caption{\textbf{The continual learning settings for Gym Control and ProcGen.} The task switches for continual CartPole \textit{(above)} and Starpilot \textit{(below)} are illustrated for demonstration.}
\vspace{-0.4cm}
\label{figure:continual_exp_envs}
\end{figure}

\vspace{-0.2cm}
\paragraph{Setups} For Gym Control, we use four environments: CartPole-v1, Acrobot-v1, LunarLander-v2 and MountainCar-v0.
For each environment, a task sequence $\mathbb{T}$ is built by chaining $k$ instances of the environment with a unique
Gaussian observation noise $\epsilon_i \sim \mathcal{N}(0, \sigma^{2})$ sampled once for each. The number $k$ is 10
for all the environments, except that $k$ is 5
for MountainCar-v0.
For ProcGen, we use all sixteen environments in the suit, while only four (i.e., Starpilot, Fruitbot, Chaser, Dodgeball) were adopted in~\citep{Muppidi2024TRAC}.
For each environment, the task sequence consists of 5 instances procedurally generated by sampling a unique game level. The budget is 2M steps per task.
Figure~\ref{figure:continual_exp_envs} illustrates the continual RL setups.

\vspace{-0.3cm}
\paragraph{Baselines} To calibrate the performance of continual RL, we make use of a standard Proximal Policy Optimization (PPO)~\citep{SchulmanWDRK17PPO} agent as the \textit{vanilla} baseline.
The agent is initialized once before the continual learning starts.
On the other side, we use a PPO agent that gets fully re-initialized every time the task is switched to be the \textit{oracle} baseline.
The oracle baseline learns each task from random initialization, thus free of the influence of task change.
Additionally, we consider six related methods including TRAC~\citep{Muppidi2024TRAC}, Weight Clipping~\citep{Elsayed2024Weightclipping} and L2 Init (also called Regenerative Regularization)~\citep{Kumar2023MaintainingPI}, Recycles Dormant Neurons (ReDo)~\citep{Sokar2302Dormant}, Layer Normalization~\citep{Lyle2024DisentanglingTC}, Adam with Relative Timesteps (AdamRel)~\citep{Ellis2024AdamOnLocalTime} as our baselines.
Note that TRAC outperforms other related methods like Concatenated ReLU~\citep{AbbasZM0M23LossofPlasticity}, EWC~\citep{Schwarz0LGTPH18EWC}, Modulating Masks~\citep{NathPBLDLKS23SharingLifelongKnowledge} in the same settings adopted in our experiments. Therefore, we do not include them here.
Besides, we found that L2 regularization performed very similarly (slightly worse) to L2 Init in our experiments, thus we use L2 Init for representation as they are very similar when the network parameterization is close to zero.

\vspace{-0.3cm}
\paragraph{Implementation} We use the official implementation of TRAC as the code base, based on which we implement all the baselines above as well as C-CHAIN (i.e., Algorithm~\ref{alg:c_chain} with the vanilla PPO agent as $\mathbb{A}$) to ensure that they only differ in the corresponding algorithm features. 
One thing to note is that C-CHAIN PPO, both $B_{\text{train}}$
and $B_{\text{ref}}$ are sampled from the online interaction data collected by the policy in the current iteration. Therefore, CHAIN PPO does not use a cross-task buffer and does not need to be aware of task switches.
For the hyperparameters of the PPO baseline agent, we use the default values and keep them consistent across all the methods.
For the hyperparameters specific to each method, we search around the recommendation values in the original papers and report the best. More experiment details are provided in Appendix~\ref{app:implementation_details}.

\begin{figure}
\begin{center}
\includegraphics[width=0.45\textwidth]{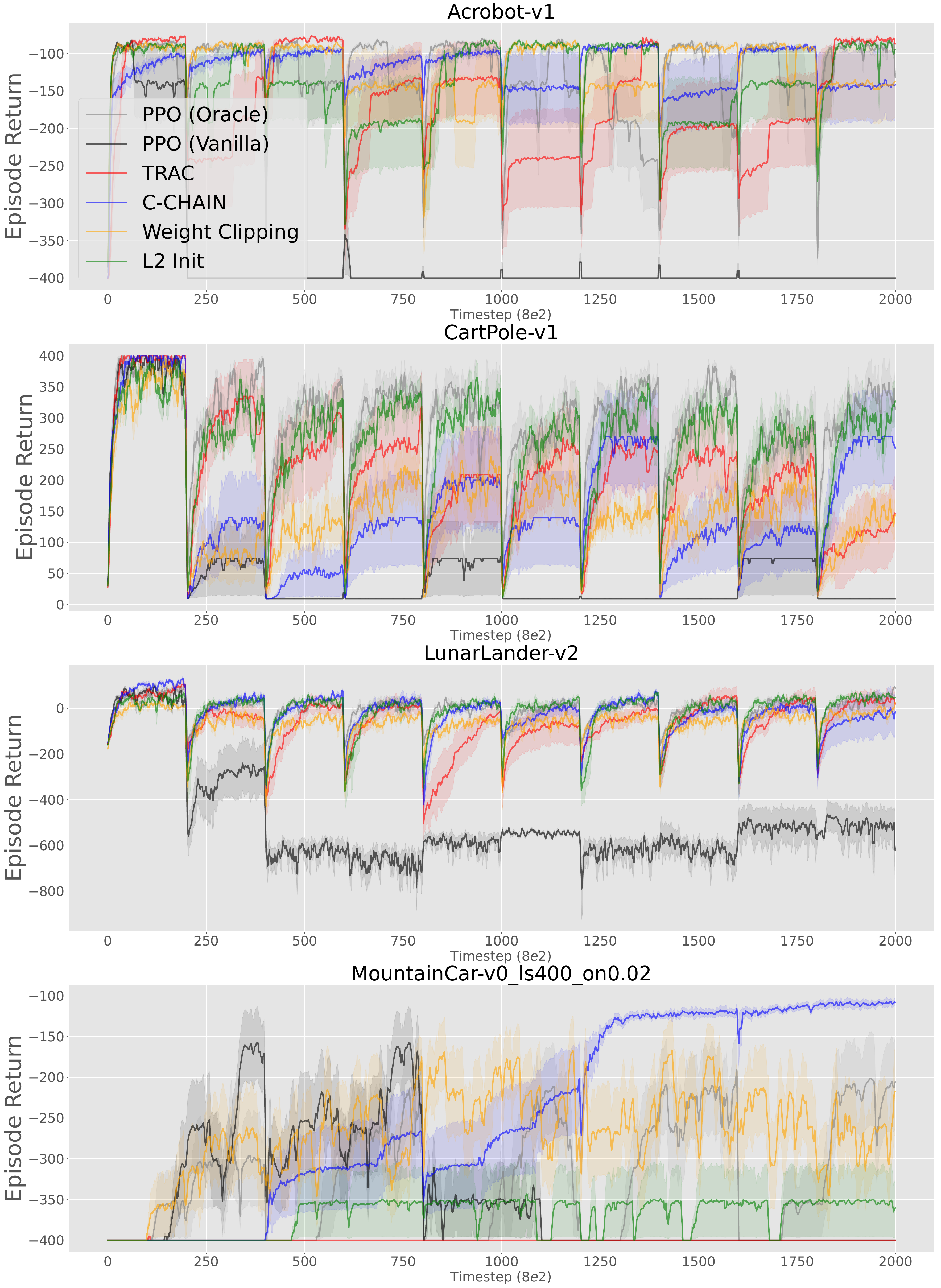}
\end{center}
\vspace{-0.6cm}
\caption{\textbf{Learning curves of different methods in four continual Gym Control environments.} The curves and the shades are means and standard errors over six seeds.}
\vspace{-0.5cm}
\label{figure:classiccontrol_eval}
\end{figure}

\begin{figure}
\begin{center}
\includegraphics[width=0.49\textwidth]{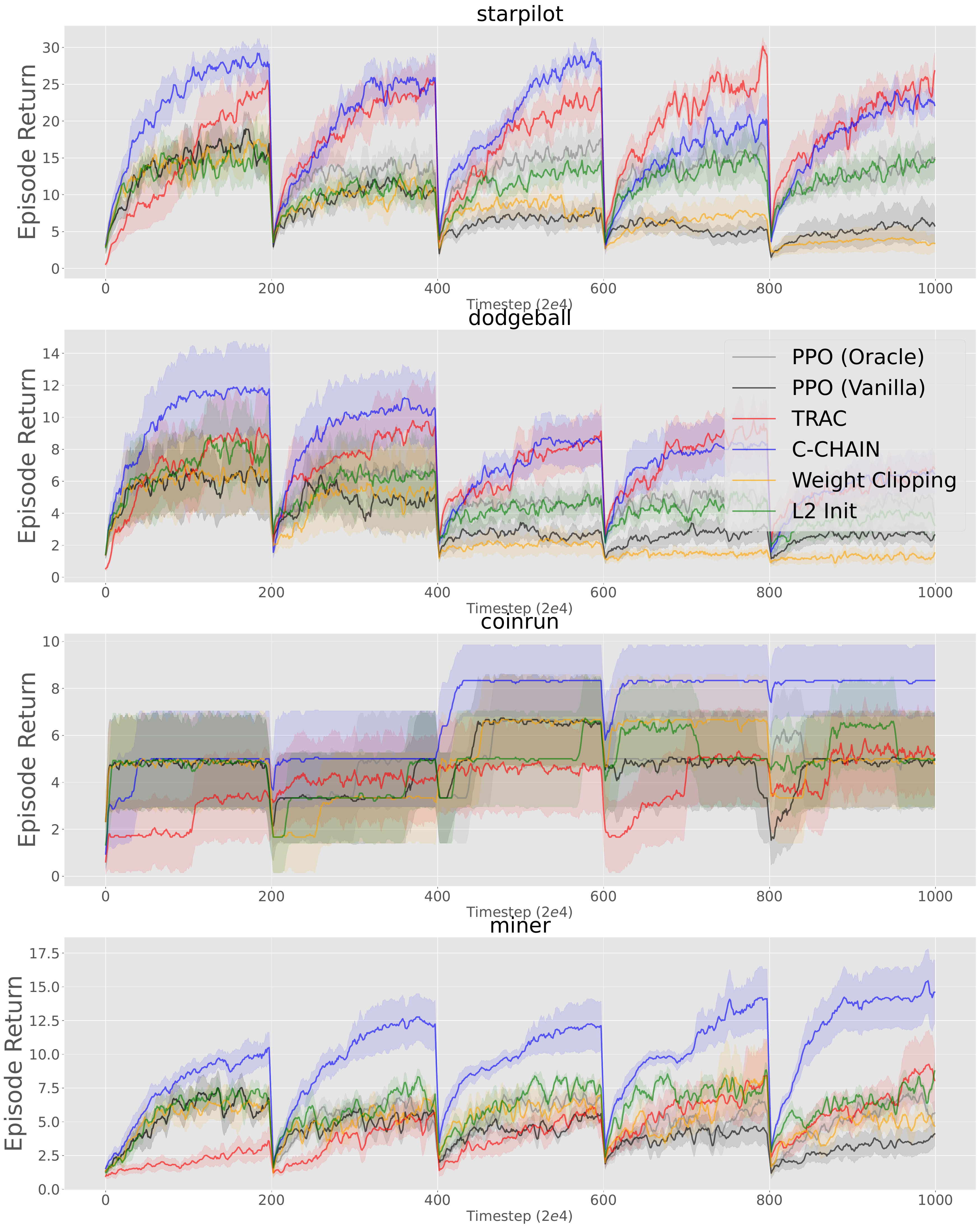}
\end{center}
\vspace{-0.6cm}
\caption{\textbf{Learning curves of different methods in four (of sixteen) continual ProcGen environments.} The curves and the shades are means and standard errors over six seeds.}
\vspace{-0.5cm}
\label{figure:procgen_eval}
\end{figure}

\vspace{-0.3cm}
\paragraph{Quantitative Comparison}
Table~\ref{table:crl_control_results_summary},\ref{table:crl_procgen_results_summary} summarizes the quantitative evaluation for different methods on continual Gym Control and ProcGen, respectively.
Figure~\ref{figure:classiccontrol_eval} and Figure~\ref{figure:procgen_eval} show the learning curves. For clarity, we plot the learning curves for three related baseline methods for ProcGen.
As task switch happens frequently, the learning curves of all methods show a phasic pattern.
For the calibration baselines, i.e., Vanilla and Oracle, Vanilla quickly degrades and collapses (especially in Gym Control) after learning the first task in the sequence, while Oracle learns each task as it re-initializes when task switch happens.
Overall, C-CHAIN and the other baseline methods effectively prevent the collapse of Vanilla and improve the learning performance greatly.
One may note that TRAC, L2 Init and C-CHAIN outperform Oracle in ProcGen. This is because, aside from task switch, non-stationarity also exists in the learning process of single-MDP RL.
Oracle still suffers from plasticity loss due to the change of sampling and policy learning within each task, while C-CHAIN also mitigates it effectively.
This also indicates that naive network resetting is not adequate to address plasticity loss in continual RL.

Among the methods in comparison, TRAC is a competitive baseline in most environments except for the failure in environments like MountainCar and Coinrun.
Weight Clipping works well in Gym Control but performs almost on par with Vanilla in ProcGen.
Similarly, L2 Init performs the best in LunarLander and CartPole while it falls below C-CHAIN by a clear margin.
A possible explanation is that both Weight Clipping and L2 Init mitigate plasticity loss by constraining the feasible parameter space near the initialization. Therefore, for problems where the agent needs a complex solution that is not easy to be found near the initialization, Weight Clipping and L2 Init can limit the learning in this sense.
Our method C-CHAIN achieves the best aggregation scores in both continual Gym Control and ProcGen, demonstrating its superiority in improving continual RL.

\begin{figure*}
\centering
\includegraphics[width=1\linewidth]{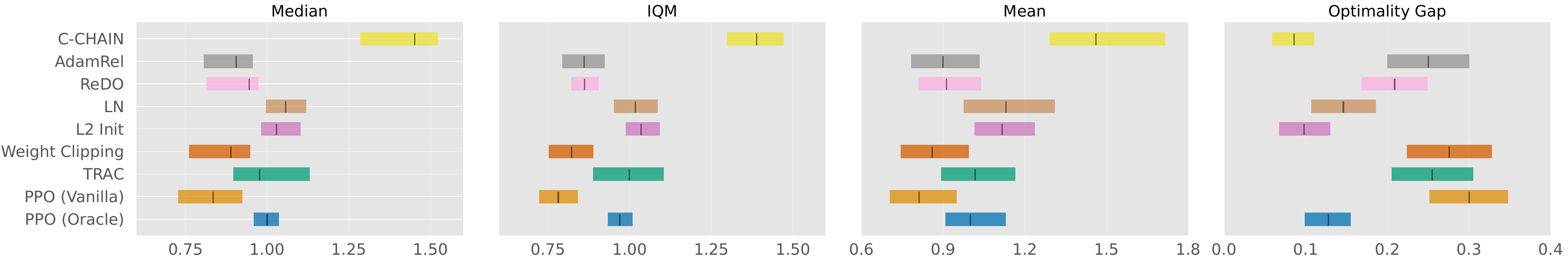}
\vspace{-0.6cm}
\caption{\textbf{Performance comparison with Reliable metrics~\citep{agarwal2021deep}.} For each method in the comparison, the results are aggregated over sixteen continual ProcGen environments with six random seeds for each.
}
\label{figure:reliable_eval_results}
\vspace{-0.2cm}
\end{figure*}

Moreover, for continual MountainCar that needs more exploration and is a bit more difficult than the other three Gym control tasks, we notice that TRAC almost totally fail on it. 
The low early-stage performance of C-CHAIN might also be due to the exploration feature of MountainCar, as reducing churn could also slow down the generalization of exploration behavior learned by the policy. Therefore, C-CHAIN learns slowly but improves steadily, while vanilla PPO learns quicker and collapses.
Continual ProcGen environments like Leaper, Jumper have sparse or even episodic rewards.
To gain more understanding for these environments, we looked into the average scores of the Max, Mean, Min curves and observed that C-CHAIN improves the average Max scores over PPO Vanilla and the average Mean scores (either by improving Max or by reducing failure numbers); while it does not fully avoid the (near-)zero Min scores due to the limited exploration ability of PPO base.

\vspace{-0.3cm}
\paragraph{Reliable Metrics}
To obtain a conclusive evaluation, we provide the aggregation evaluation by using Reliable metrics~\cite{agarwal2021deep}. The evaluation uses Mean, Median, Inter-Quantile Mean (IQM) and Optimality Gap, with 95\% confidence intervals. It aggregates over 9 methods, 6 seeds for each of 16 ProcGen tasks as reported in Table~\ref{table:crl_procgen_results_summary}, i.e., 864 runs in total.

The results are shown in Figure~\ref{figure:reliable_eval_results}.
We can observe that C-CHAIN performs the best regarding all four metrics and outperforms the second-best with no overlap of Confidence Intervals for Median, IQM, and a minor one for Mean.

\begin{figure}
\begin{center}
\includegraphics[width=0.45\textwidth]{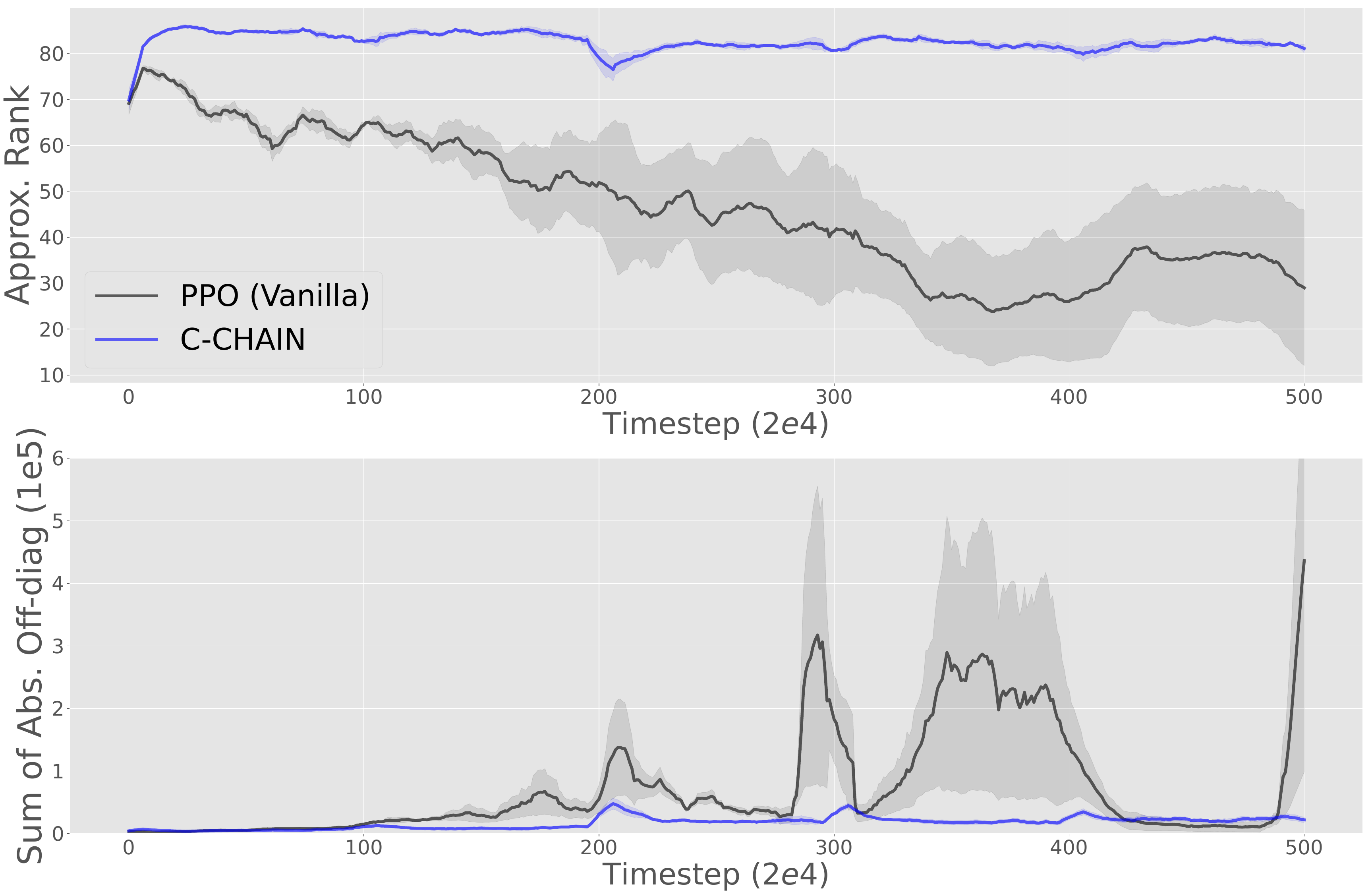}
\end{center}
\vspace{-0.6cm}
\caption{\textbf{Analysis on NTK matrix} in terms of approximate rank \textit{(above)} and the sum of absolute off-diagonal values \textit{(below)} in continual ProcGen Starpilot.}
\label{figure:procgen_ntk_stats}
\vspace{-0.4cm}
\end{figure}

\subsection{Empirical Analysis}
\label{subsec:empirical_analysis}

To better understanding how C-CHAIN improves continual RL and examine whether the empirical efficacy of C-CHAIN matches our formal analysis.
We conduct empirical analysis on (1) the rank evolvement of the empirical NTK matrix (corresponding to Section~\ref{subsec:learning_dynamics}) and (2) the ablation of the two effects of reducing churn (i.e., \ding{172}, \ding{173} in Section~\ref{subsec:dissecting_the_effect_of_churn_reduction}).

\vspace{-0.2cm}
\paragraph{On the Evolvement of the Empirical NTK Matrix}

We empirically compute and record the NTK matrix of the agent's policy network at every iteration for the analysis.
The details are provided in Appendix~\ref{app:ntk_analysis_implementation_detal}.
Figure~\ref{figure:procgen_ntk_stats} shows the approximate rank~\citep{KumarAGL21ImplicitUnder} of the empirical NTK matrix and the summation of absolute off-diagonal values.
For the vanilla PPO agent, the curve of approximate rank exhibits large instability, overall decreasing to a low rank throughout learning.
This corresponds to the sharp increase of the scale of off-diagonal values.
As expected, C-CHAIN suppresses the off-diagonal values of the NTK matrix and maintains a stable and high rank.
This explains the performance of C-CHAIN in Figure~\ref{figure:procgen_eval} and Table~\ref{table:crl_control_results_summary},\ref{table:crl_procgen_results_summary}, and empirically confirms our formal analysis in Section~\ref{sec:method}.

\begin{figure}
\begin{center}
\includegraphics[width=0.2085\textwidth]{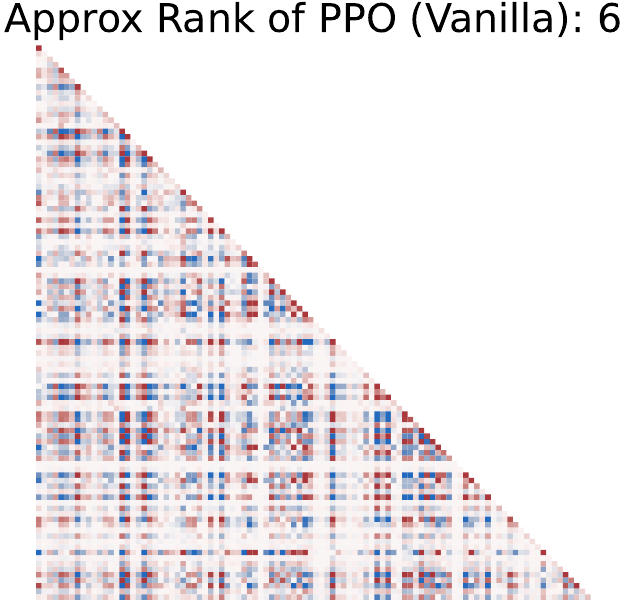}
\hspace{0.1cm}
\includegraphics[width=0.23\textwidth]{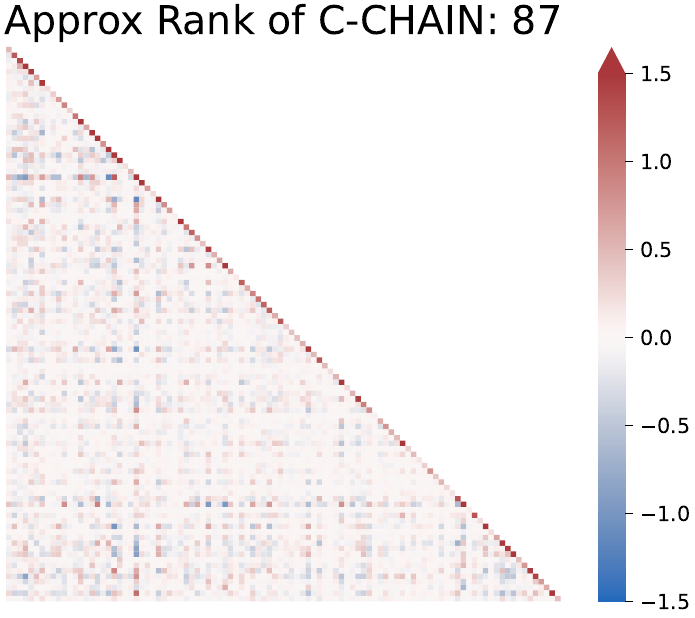}
\end{center}
\vspace{-0.5cm}
\caption{\textbf{Visualization of the empirical NTK matrix (100 by 100)} for Vanilla \textit{(left)} and C-CHAIN \textit{(right)} at 8M timesteps (10M in total) in continual ProcGen Starpilot.}
\vspace{-0.3cm}
\label{figure:procgen_ntk_visual}
\end{figure}

Moreover, as shown by the visualization in Figure~\ref{figure:procgen_ntk_visual}, we can observe that the vanilla PPO agent exhibits a highly correlated NTK with off-diagonal cells of large absolute values (i.e., blue or red). The existence of high correlation leads to low rank, i.e., the sign of plasticity loss, which also echoes the empirical findings in~\citep{Lyle2024DisentanglingTC}.
In contrast, C-CHAIN shows a less correlated NTK and thus maintains the plasticity of the agent.

\vspace{-0.2cm}
\paragraph{On the Ablation of the Two Effects of C-CHAIN}
We empirically dissect the gradient (denoted by $\bar g^{\text{cr}}$) of C-CHAIN's churn reduction loss function (Equation~\ref{eq:churn_reduction_loss}) into a projective component (i.e., $\textbf{proj}_{g^{\text{PPO}}} \bar g^{\text{cr}}$ and $g^{\text{PPO}}$ for the gradient of the PPO base agent) and an orthogonal component (i.e., $g^{\text{cr}} - \textbf{proj}_{g^{\text{PPO}}} \bar g^{\text{cr}}$).
We use the projective component to approximate the \ding{173} term and use the orthogonal component for the \ding{172} term.
By applying either component only, we have two variants of C-CHAIN, denoted by \textit{Proj Only} and \textit{Orth Only}.
We then evaluate the contribution of the two components by comparing with C-CHAIN and Vanilla.

\begin{figure}
\begin{center}
\includegraphics[width=0.44\textwidth]{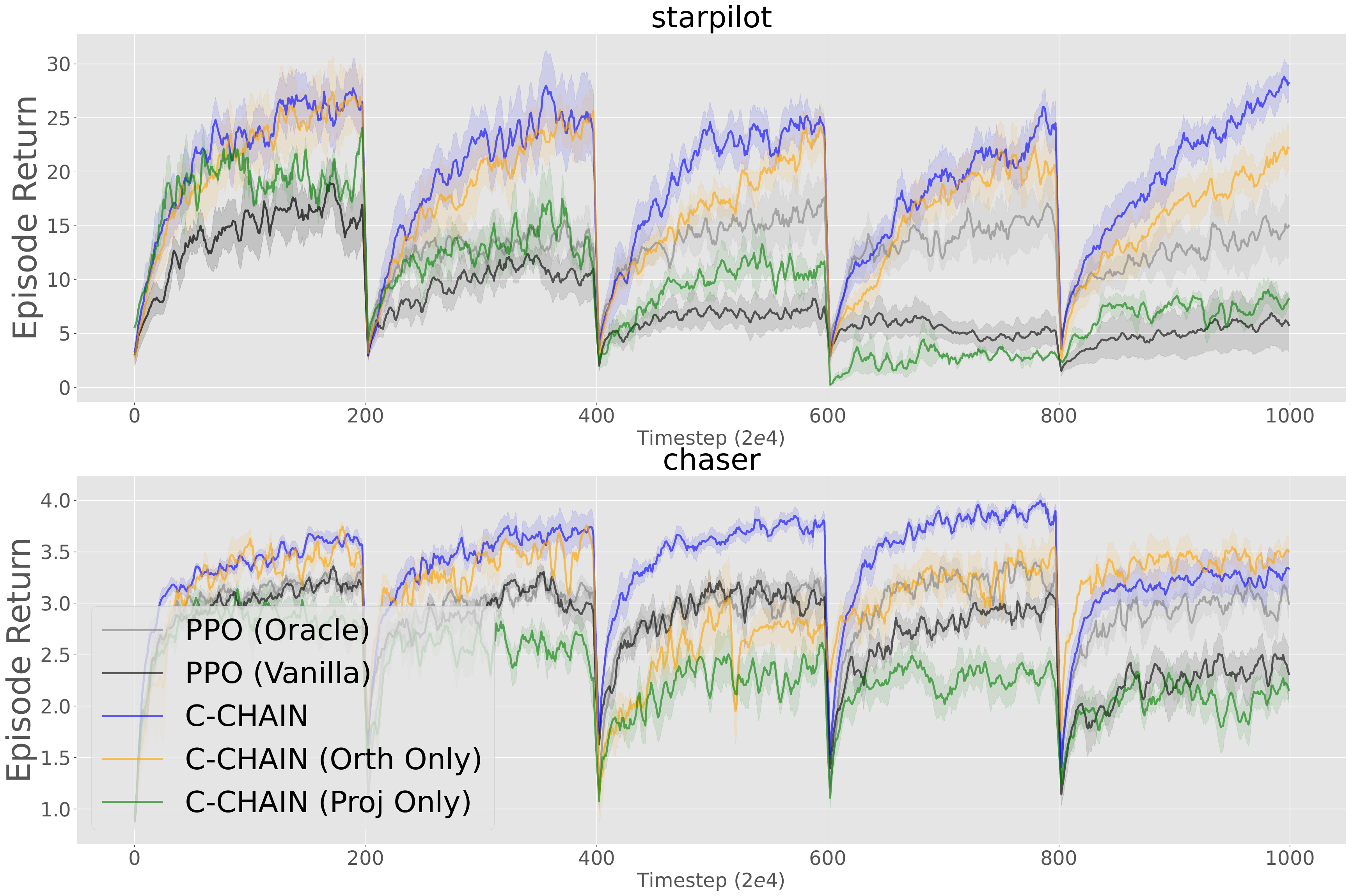}
\end{center}
\vspace{-0.6cm}
\caption{\textbf{Ablation study for the projective and orthogonal gradient components} of C-CHAIN in ProcGen Starpilot and Chaser.}
\vspace{-0.5cm}
\label{figure:procgen_proj}
\end{figure}

As in Figure~\ref{figure:procgen_proj}, we can observe that both the orthogonal and the projective components are beneficial to the vanilla baseline.
The two components contribute collectively and add up to the effectiveness of C-CHAIN.
Moreover, the orthogonal component contributes more than the projective component.
This indicates that suppressing the off-diagonal entries of the NTK (i.e., the \ding{172} term) is critical to C-CHAIN.
This naturally delineates the difference between the efficacy of C-CHAIN and the related methods that alters RL gradient with projective information in different ways~\citep{ChaudhryRRE19AGEM,YuK0LHF20GradientSurgery}.

\subsection{C-CHAIN in More Continual Learning Settings}
\label{subsec:more_results}

\paragraph{Continual DeepMind Control Suite (DMC)} In addition to the discrete-action settings above, we evaluate C-CHAIN in continual continuous control. We build three continual RL environments based on DMC~\citep{Tassa2018DMC}: \textit{Continual Walker} (Stand-Walk-Run), \textit{Continual Quadruped} (Walk-Run-Walk), \textit{Continual Dog} (Stand-Walk-Run-Trot), each of which is a continual RL scenario that chains the corresponding individual tasks in a sequence.
\begin{table}[h]
  \vspace{-0.6cm}
  \caption{\textbf{Performance comparison in continual DMC.} The reported scores are means and standard errors over twelve seeds.}
  \vspace{0.1cm}
  \centering
  \scalebox{0.68}{
  \begin{tabular}{c|c|c|c}
    \toprule
    Env & Oracle & Vanilla & C-CHAIN \\
    \midrule
    C-Walker & 395.971 $\pm$ 8.116 & 305.199 $\pm$ 18.519 & \highlightbf{472.828 \pm 17.865} \\
    C-Quadruped & 234.529 $\pm$ 19.193 & 250.153 $\pm$ 26.385 & \highlightbf{314.510 \pm 44.321} \\
    C-Dog & 137.080 $\pm$ 3.133 & 129.744 $\pm$ 5.914 & \highlightbf{174.098 \pm 9.169} \\
    \bottomrule
  \end{tabular}
  }
\label{table:crl_dmc_results}
\vspace{-0.5cm}
\end{table}

The results are shown in Table~\ref{table:crl_dmc_results}.
We can observe that C-CHAIN PPO significantly outperforms PPO in all three settings, showing the effectiveness for continual control. C-CHAIN also outperforms PPO Oracle because it does not reset for each task and thus can \textit{transfer} the learned skill from previous tasks to future ones (e.g., Dog-Stand to Dog-Walk). Details are provided in Appendix~\ref{app:continual_dmc_details}.

\vspace{-0.3cm}
\paragraph{Continual MinAtar}
To evaluate the generality across different base agents, we implement C-CHAIN DoubleDQN~\citep{HasseltGS16DoubleDQN} and evaluate it in a continual MinAtar~\citep{Young2019MinAtar} setting built by us. Specifically, we chain three tasks: SpaceInvaders, Asterix, Seaquest.
The results in Table~\ref{table:crl_minatar_results} show that C-CHAIN also improves the continual learning performance upon DoubleDQN in a sequence of totally \textit{different} tasks. Details are provided in Appendix~\ref{app:continual_mintar_details}.

\begin{table}[h]
  \vspace{-0.6cm}
  \caption{\textbf{Performance comparison in continual MinAtar.} The reported scores are means and standard errors over twelve seeds.}
  \vspace{0.1cm}
  \centering
  \scalebox{0.68}{
  \begin{tabular}{c|c|c}
    \toprule
    Env & Vanilla & C-CHAIN \\
    \midrule
    C-MinAtar & 22.044 $\pm$ 0.733 & \highlightbf{29.513 \pm 0.682} \\
    \bottomrule
  \end{tabular}
  }
\label{table:crl_minatar_results}
\vspace{-0.3cm}
\end{table}

\vspace{-0.3cm}
\paragraph{Continual Supervised Learning (SL) on MNIST} We extend our empirical evaluation of C-CHAIN to continual supervised learning setting.
We follow the settings in L2 Init~\citep{Kumar2023MaintainingPI} and adopt RandomLabel-MNIST and Permuted-MNIST as our testbed.
The results and implementation details can be found in Table~\ref{table:continual_supervised_learning} and Appendix~\ref{app:continual_supervised_learning_details}.

The results show that C-CHAIN improves the vanilla agent but does not perform on par with L2 Init and Weight Clipping.
The efficacy of C-CHAIN is relatively limited in the two continual SL settings, in contrast to its superiority in the continual RL experiments.
We hypothesize this is because RL suffers more from churn accumulation due to the chain effect~\citep{Tang2024ReduceChurn}, which acts as a prominent cause of plasticity loss in continual RL, thus C-CHAIN can address it effectively.
Besides, the two MNIST settings are simple, where a good solution can be found near the initialization as preferred by L2 Init and Weight Clipping. A more thorough study on this point is expected in the future with more continual supervised learning settings.

\section{Conclusion}
\label{sec:conclusion}

In this paper, we study plasticity loss from the lens of churn. We established a formal connection between the two with NTK and demonstrated the interplay between the rank decrease of NTK and 
the exacerbation of churn in learning dynamics.
We show that our method C-CHAIN effectively improves continual RL results across a wide range of tasks.

\vspace{-0.3cm}
\paragraph{Limitation and Future work}

Beyond theoretical advancements, applying churn reduction to real-world continual learning settings, as robotics or large language models (LLMs), could address plasticity loss and catastrophic forgetting in adaptive learning settings \citep{wu2024continual,zhai2024investigating}. Investigating its role in reinforcement learning from human feedback (RLHF) could improve stability in policy updates, ensuring LLMs retain learned behaviors while adapting to evolving human preferences. 
Besides, Equation~\ref{eq:the_first_term} reveals an interesting direction to realizing churn reduction in a more direct and effective way by estimating Hessian~\citep{Elsayed2024RevisitingHessian}.

\clearpage

\section*{Acknowledgements}
We want to acknowledge funding support from NSERC, FQRNT, Google, CIFAR AI and compute support from Digital Research Alliance of Canada, Mila IDT, and NVidia.

We thank the anonymous reviewers for their valuable help in improving our manuscript. We would also like to thank the Python community \cite{van1995python, 4160250} for developing tools that enabled this work, including NumPy \cite{harris2020array}, Matplotlib \cite{hunter2007matplotlib}, Jupyter \cite{2016ppap} and Pandas \cite{McKinney2013Python}.

\section*{Impact Statement}

This paper presents work whose goal is to advance the field of 
Machine Learning. There are many potential societal consequences 
of our work, none which we feel must be specifically highlighted here.

\bibliography{chain_crl}
\bibliographystyle{icml2025}

\newpage
\appendix
\onecolumn

\section{Experimental Details}
\label{app:implementation_details}

\subsection{Continual RL}
\label{app:crl_details}

\paragraph{Environment Setups}
We follow the setups in~\citep{Muppidi2024TRAC}.
For Gym Control, we use four environments:  CartPole-v1, Acrobot-v1, LunarLander-v2 and MountainCar-v0.
For each environment, a task sequence $\mathbb{T}$ is built by chaining $k$ instances of the environment with a unique
Gaussian observation noise $\epsilon_i \sim \mathcal{N}(0, \sigma^{2})$ sampled once for each.
The number $k$ is 10 and the interaction budget $N$ is 0.16M for all the environments, except that $k$ is 5 and $N$ is 0.32M for MountainCar-v0.
For CartPole-v1, Acrobot-v1 and LunarLander-v2, we use $\sigma = 2.0$ as provided in~\citep{Muppidi2024TRAC}.
For MountainCar-v0, it was not originally included in~\citep{Muppidi2024TRAC} and we found $\sigma = 2.0$ is too large for this environment to be learnable. Therefore, we use $\sigma = 0.02$ for MountainCar-v0.

For ProcGen, we use all sixteen environments in the suite, while only four (i.e., Starpilot, Fruitbot, Chaser, Dodgeball) were adopted in~\citep{Muppidi2024TRAC}.
For each environment, the task sequence consists of 5 instances procedurally generated by sampling a unique level of the environment. The budget is 2M steps per task.

\paragraph{Algorithm Implementation}

We use the official implementation\footnote{\url{https://github.com/ComputationalRobotics/TRAC}} of TRAC~\citep{Muppidi2024TRAC} as our code base.
We implement all other methods to
ensure that they only differ in the corresponding algorithm
features. For the hyperparameters of the PPO baseline agent,
we use the default values recommended in the code base and keep them consistent across
all the methods. For the hyperparameters specific to each
method, we search around the recommendation values in
the original papers and report the best.

For C-CHAIN, we follow the practice in~\citep{Tang2024ReduceChurn} and we only reduce churn for the policy network of the PPO agent.
The coefficient $\lambda_{\pi}$ of the policy churn reduction regularization is determined by an auto-adjustment mechanism, which keeps a consistent relative scale (denoted by $\beta$) between the churn reduction regularization terms and the original DRL objectives.
Specifically, by maintaining the running means of the absolute PPO loss $| \bar L_{\pi}|$ and the policy churn reduction term $| \bar  L_{\text{PC}} | $, the churn reduction regularization coefficient $\lambda_{\pi}$ is computed dynamically as $\lambda_{\pi} = \beta \frac{| \bar L_{\pi}| }{| \bar  L_{\text{PC}} | }$.
For the size of $B_{\text{ref}}$, we did the experiments for different batch sizes for on continual Gym control tasks. Our findings were that using a 2x, 4x or 8x batch size for $B_{\text{ref}}$ (compared to the regular training batch size) sometimes improved the learning performance, but not consistently. To some degree, increasing the batch size of $B_{\text{ref}}$ acted similarly to increasing the regularization coefficient, as both of them reduce more churn. Thus, we did not search for the best batch size for $B_{\text{ref}}$ and set it equal to the training batch size to alleviate the hyperparameter choice burden.

The hyperparameters are provided in Table~\ref{table:hyperparams}.
The code will be released when this work is made public.

\begin{table}[h]
  \caption{\textbf{Hyperparameters of different methods used in continual Gym Control and ProcGen environments.} 
  We use `-' to denote the `not applicable' situation.
  }
  \vspace{0.2cm}
  \centering
  \scalebox{0.8}{
  \begin{tabular}{c|c|c|c}
    \toprule
    Agent & Hyperparam & Gym Control & ProcGen \\
    \midrule
    \multirow{9}{*}{Vanilla PPO~\citep{SchulmanWDRK17PPO}} & Learning Rate & 1e$^{-3}$ & 1e$^{-3}$\\
    & Optimizer & Adam & Adam\\
    & Discount Factor ($\gamma$) & 0.99 & 0.99 \\
    & GAE Parameter ($\lambda$) & 0.95 & 0.95 \\
    & Training Interval & 800 steps & 1000 steps \\
    & Mini-batch Size & 32 & 125\\
    & Update Epoch & 5 & 3 \\
    & Clipping Range Parameter ($\epsilon$)
     & 0.2 & 0.2 \\
    & Entropy Loss Coef 
     & 0 & 0.01 \\
    \midrule
    TRAC~\citep{Muppidi2024TRAC} & (Hyperparam Free) & - & - \\
    \midrule
    Weight Clipping~\citep{Elsayed2024Weightclipping} & Clipping Range & Best in $\{0.1, 0.5, 1, 10\}$ & Best in $\{0.1, 0.5, 1, 10\}$ \\
    \midrule
    L2 Init~\citep{Kumar2023MaintainingPI} & Regularization Coeff & Best in $\{0.1, 1, 10, 100, 1000\}$ &  Best in $\{0.1, 1, 10, 100, 1000\}$\\
    \midrule
    C-CHAIN (Ours) & Target Relative Loss $\beta$ for Auto $\lambda_{\pi}$
     & Best in $\{1000, 1e4, 1e5\}$ & Best in $\{1000, 1e4, 1e5\}$ \\
    \bottomrule
  \end{tabular}
  }
\label{table:hyperparams}
\end{table}

\paragraph{Compute Resource}

For the continual Gym Control and ProcGen experiments, we allocate a single V100 GPU, 16 CPUs and 32GB memory for 4 to 6 jobs, typically running for around 4 hours for Gym Control and 20 hours for ProcGen to complete.

\subsection{Continual DeepMind Control Suite}
\label{app:continual_dmc_details}

We use the public implementations of PPO and DMC task setups in \texttt{CleanRL}\footnote{\url{https://github.com/vwxyzjn/cleanrl}} as our codebase.
The actor and critic networks are two-layer MLPs with 256 units for each layer.
We made no change to the network structure, recommended hyperparameter choices, etc. The hyperparameters are in Table~\ref{table:ppo_dmc_hyperparameter}.

Upon the code base, we build three continual RL settings in with DMC tasks:
\begin{itemize}
    \item \textit{Continual Walker}: run Walker-stand, Walker-walk, Walker-run, sequentially.
    \item \textit{Continual Quadruped}: run Quadruped-walk, Quadruped-run, Quadruped-walk, sequentially (repeat because only two Quadruped tasks are available in DMC).
    \item \textit{Continual Dog}: run Dog-stand, Dog-walk, Dog-run, Dog-trot, sequentially.
\end{itemize}
We run 1M steps for each task.

\begin{table}[ht]
\vspace{-0.3cm}
  \caption{Hyperparameters of PPO and C-CHAIN used in continual DMC environments. The values of conventional hyperparameters are taken from the recommended values in \texttt{CleanRL}.
  }
  \vspace{0.1cm}
  \centering
  \scalebox{0.95}{
  \begin{tabular}{c|c}
    \toprule
    \multicolumn{2}{c}{PPO Hyperparameters}\\
    \midrule
    Learning Rate & 3e$^{-4}$\\
    Training Interval & 2048 steps \\
    Discount Factor ($\gamma$) & 0.99 \\
    GAE Parameter ($\lambda$) & 0.95 \\
    Num. of Minibatches & 32 \\
    Update Epoch & 10 \\
    Clipping Range Parameter ($\epsilon$)
    & 0.2 \\
    \midrule
    \multirow{2}{*}{Target Relative Loss $\beta$ for Auto $\lambda_{\pi}$}
     & 0.5 for Continual Walker \\
    & 0.05 for Continual Quadruped and Continual Dog \\
    \bottomrule
  \end{tabular}
  }
\label{table:ppo_dmc_hyperparameter}
\vspace{-0.2cm}
\end{table}

\begin{figure*}
\vspace{-0.2cm}
\centering
\subfigure[Continual Walker]{
\includegraphics[width=0.7\textwidth]{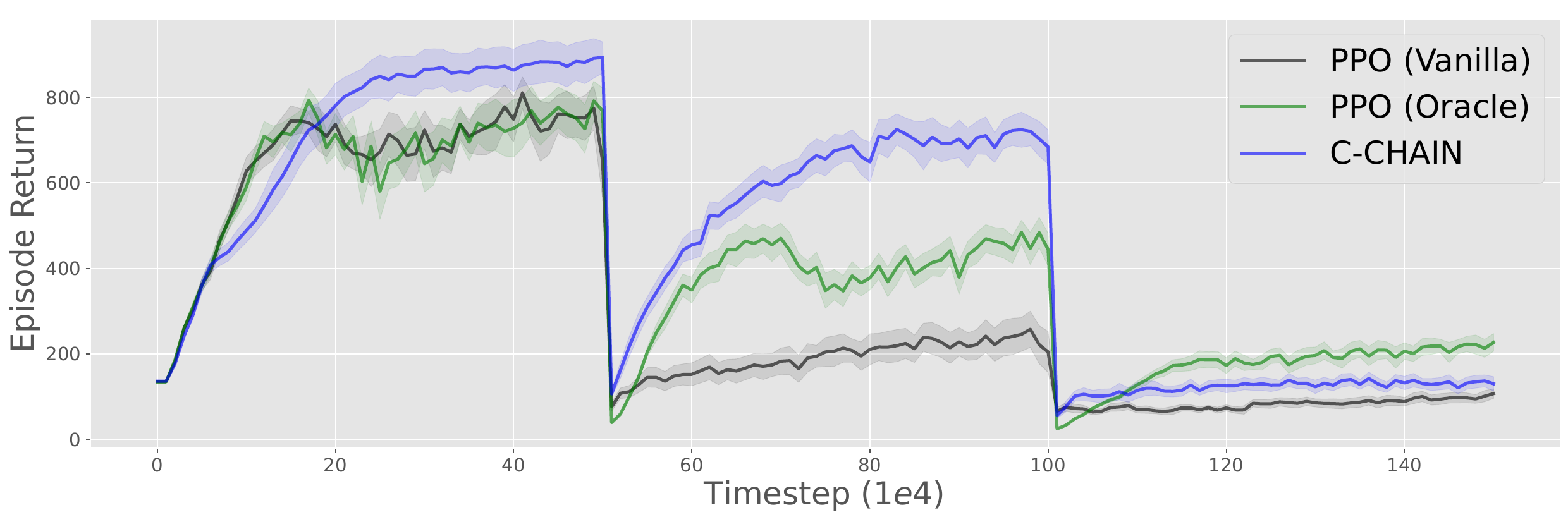}
}
\vspace{-0.2cm}
\subfigure[Continual Quadruped]{
\includegraphics[width=0.7\textwidth]{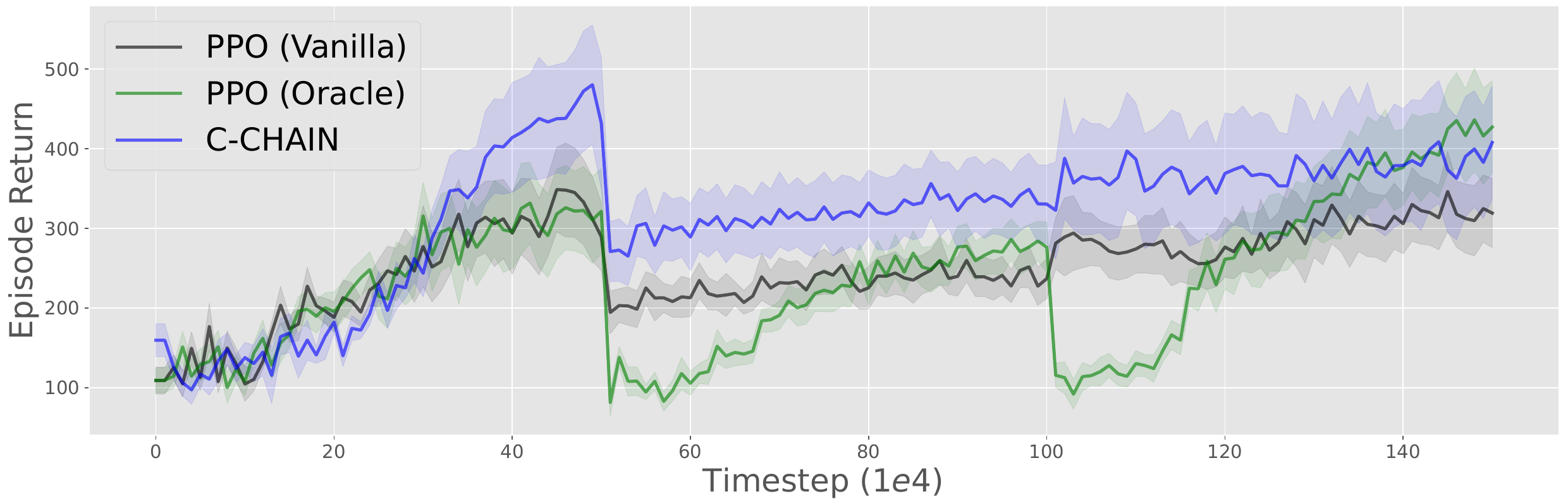}
}
\vspace{-0.2cm}
\subfigure[Continual Dog]{
\includegraphics[width=0.7\textwidth]{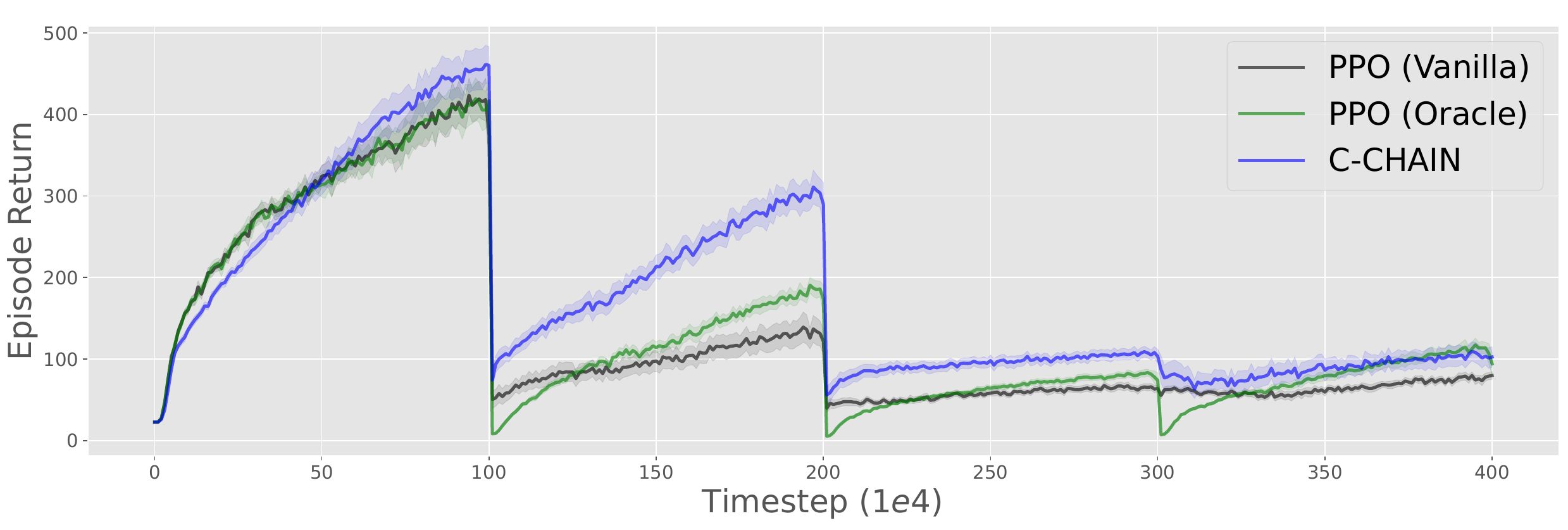}
}
\vspace{-0.2cm}
\caption{
\textbf{Learning curves for PPO and C-CHAIN PPO in continual DMC.} The curves and the shades are means and standard errors over twelve seeds.
}
\label{figure:continual_dmc_curves}
\end{figure*}

Similarly, we compare PPO Oracle, PPO Vanilla, and C-CHAIN in the three settings. The results are summarized in Table~\ref{table:crl_dmc_results} and Figure~\ref{figure:continual_dmc_curves} shows the learning curves.
We can observe that C-CHAIN PPO significantly outperforms PPO in all three settings, showing the effectiveness for continual control. C-CHAIN also outperforms PPO Oracle because it does not reset for each task and thus can \textit{transfer} the learned skill from previous tasks to future ones (e.g., Dog-Stand to Dog-Walk).

\subsection{Continual MinAtar}
\label{app:continual_mintar_details}

We use the official code of MinAtar paper\footnote{\url{https://github.com/kenjyoung/MinAtar}} as our codebase.
For DoubleDQN, we modify the DQN implementation provided in the official MinAtar code with no change to the network structure, recommended hyperparameter choices, etc.
We use DoubleDQN as the base agent, and apply C-CHAIN to the training of the value network, with no modification to the hyperparameters of DoubleDQN.
The hyperparameters are summarized in Table~\ref{table:minatar_hyperparm_common}.

\begin{table}[ht]
\vspace{-0.3cm}
  \caption{Hyperparameters of DoubleDQN and C-CHAIN used in continual MinAtar environments. The values of conventional hyperparameters are taken from the recommended values  in~\citep{Young2019MinAtar}.
  }
  \vspace{0.1cm}
  \centering
  \scalebox{0.95}{
  \begin{tabular}{c|c}
    \toprule
    \multicolumn{2}{c}{DoubleDQN Hyperparameters}\\
    \midrule
    Learning Rate & 3e$^{-4}$\\
    Training Interval & 1 step \\
    Discount Factor ($\gamma$) & 0.99 \\
    Hard Replacement Interval & 1000 steps \\
    Replay Buffer Size & 0.5M \\
    Batch Size & 32 \\
    Initial $\epsilon$ & 1.0 \\
    End $\epsilon$ & 0.1 \\
    $\epsilon$ Decay Steps & 0.5M \\
    Initial Random Steps & 10k \\
    \midrule
    Target Relative Loss $\beta$ for Auto $\lambda_Q$ & 0.01 \\
    \bottomrule
  \end{tabular}
  }
\label{table:minatar_hyperparm_common}
\vspace{-0.3cm}
\end{table}

\begin{figure*}[h]
\begin{center}
\includegraphics[width=0.7\textwidth]{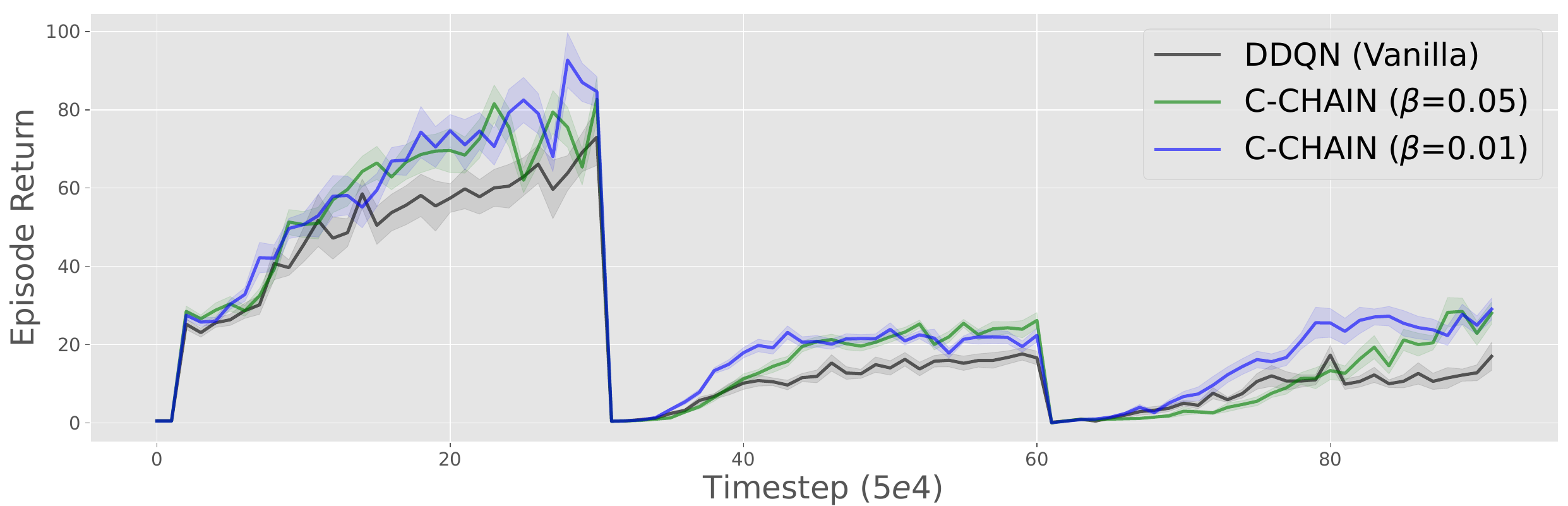}
\end{center}
\vspace{-0.6cm}
\caption{\textbf{Learning curves for DoubleDQN and C-CHAIN in continual MinAtar.} The curves and the shades are means and standard errors over twelve seeds.}
\vspace{-0.2cm}
\label{figure:continual_minatar_curves}
\end{figure*}

To build the continual MinAtar setting, we chain three tasks: SpaceInvaders, Asterix, Seaquest, by padding the observation space to be $[10, 10, 10]$. We run 1.5M steps for each task, i.e., 4.5M in total.

We compare DoubleDQN (i.e., Vanilla) and C-CHAIN DoubleDQN with 12 random seeds. The results are shown in Table~\ref{table:crl_minatar_results} and Figure~\ref{figure:continual_minatar_curves} shows the learning curves.
The results show that C-CHAIN also improves the continual learning performance upon DoubleDQN in a sequence of totally different tasks. This indicates that C-CHAIN is not only beneficial to PPO in continual RL but has the potential to be a general and easy-to-use remedy to different off-the-shelf RL algorithms in combating plasticity loss in continual RL scenarios.

\subsection{Continual Supervised Learning}
\label{app:continual_supervised_learning_details}

We follow the settings in L2 Init~\citep{Kumar2023MaintainingPI} and adopt RandomLabel-MNIST and Permuted-MNIST as our testbed.
For RandomLabel-MNIST, each task in the task sequence is a classification learning task with 1200 images randomly sampled from the MNIST dataset and random labels assigned to each individual image.
The task sequence consists of 50 tasks and we use train the agent for 400 epochs per task.
For Permuted-MNIST, each task is a classification learning task with 10000 images sampled from MNIST. Different from RandomLabel-MNIST, a Permuted MNIST task is characterized by applying a fixed randomly sampled permutation to the input pixels of all the images. The task sequence consists of 500 tasks and the agent is trained for one epoch per task.

We use the official implementation of L2 Init\footnote{\url{https://github.com/skumar9876/L2_Init}} as the code base and implement C-CHAIN and Weight Clipping with simple modifications.
We make no modifications to common hyperparameters and use the recommended values in the code base.

The results are summarized in Table~\ref{table:continual_supervised_learning}.
We can observe that C-CHAIN improves the vanilla agent but does not perform on par with L2 Init and Weight Clipping.
This shows that the efficacy of C-CHAIN is relatively limited in the two continual supervised learning environments, especially in contrast to its superiority in the continual RL experiments.
We hypothesize this is because RL suffers more from churn accumulation and thus more severe plasticity loss due to the chain effect of churn~\citep{Tang2024ReduceChurn}.
Besides, Permuted-MNIST and RandomLabel-MNIST are simple, where the agent can find a good solution near the initialization.
We expect to evaluate C-CHAIN in more complex continual supervised learning environments in the future.

\begin{table}[h]
  \vspace{-0.3cm}
  \caption{\textbf{Performance comparison in continual supervised learning environments.} The reported scores are the average performance (Equation~\ref{eq:crl_avg_performance_objective}) in terms of accuracy with mean and standard error over three seeds.}
  \vspace{0.1cm}
  \centering
  \scalebox{0.9}{
  \begin{tabular}{c|c|c}
    \toprule
    Algorithm & RandomLabel-MNIST & Permuted-MNIST \\
    \midrule
    Vanilla & 0.1501 $\pm$  0.0030 & 0.6430 $\pm$  0.0029 \\
    C-CHAIN & 0.2482 $\pm$ 0.0141 & 0.6797 $\pm$  0.0042 \\
    L2 Init & 0.8607 $\pm$  0.0021 & 0.8039 $\pm$  0.0027 \\
    Weight Clipping & 0.4304 $\pm$  0.0052 & 0.8281 $\pm$  0.0001 \\
    \bottomrule
  \end{tabular}
  }
\label{table:continual_supervised_learning}
\end{table}

\section{Empirical NTK Analysis Details}
\label{app:ntk_analysis_implementation_detal}

\begin{figure}
\begin{center}
\includegraphics[width=0.8\textwidth]{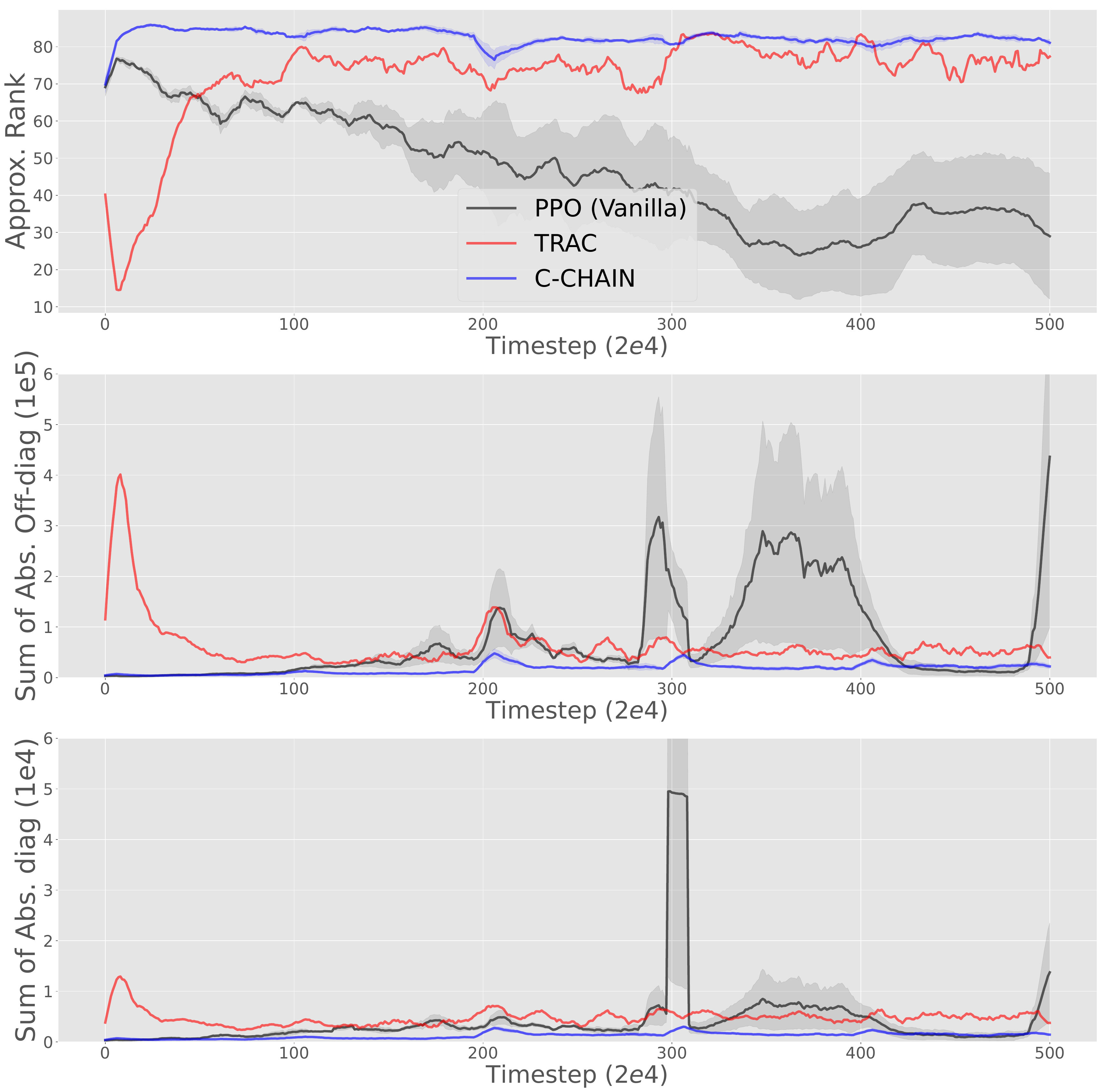}
\end{center}
\vspace{-0.3cm}
\caption{\textbf{Analysis on NTK matrix} in terms of approximate rank \textit{(above)}, the sum of absolute off-diagonal values \textit{(middle)}, and the sum of absolute diagonal values\textit{(below)} in continual ProcGen Starpilot.}
\vspace{-0.4cm}
\label{figure:procgen_ntk_stats_appendix}
\end{figure}

During the learning process, we collect the empirical NTK matrices in each training iteration of the PPO agent.
Specifically, for the interaction experience collected within each iteration, we divide it into $m$ mini-batches randomly ($m = 100$ for our NTK analysis in ProcGen).
Then, for each mini-batch, we compute the gradient of the policy network's parameters $\theta$ regarding the PPO policy optimization objective, denoted by $g_i$ with $i \in \{1, 2, \dots, m\}$.
The empirical NTK matrix $N_{\theta}$ is then computed by following the definition in Equation~\ref{eq:ntk_def}:
$N_{\theta}(i,j) = g_i^\top g_j$ for $i,j \in \{1, 2, \dots, m\}$.
With the empirical NTK matrix, statistics for the off-diagonal values (i.e., Figure~\ref{figure:procgen_ntk_stats}, \textit{below} and Figure~\ref{figure:procgen_ntk_stats_appendix}, \textit{middle}) and the diagonal values (i.e., Figure~\ref{figure:procgen_ntk_stats_appendix}, \textit{below}) can be computed conveniently.

For the computation of approximate rank, we follow the prior convention in~\citep{YangZXK20SVRL,KumarAGL21ImplicitUnder}.
For a matrix $N$, let $\{\sigma_{i}(N)\}$ be the singular values of it arranged in decreasing order, i.e., $\sigma_1 \ge \dots \ge \sigma_d \ge 0$.
the approximate rank regarding a threshold $\delta$ is defined as srank$_{\delta}(N) = \min \{k:\sum_{i=1}^{k}\sigma_{i}(N)\ge (1-\delta)\sum_{i=1}^{d}\sigma_{i}(N)\}$.
In this paper, we use $\delta=0.01$.
In other words, this means the approximate rank is the first $k$
singular values that hold more than 99\% information of all singular values.
Accordingly, we use this computation for the results of approximate rank in Figure~\ref{figure:procgen_ntk_stats} \textit{(above)} and Figure~\ref{figure:procgen_ntk_stats_appendix} \textit{(above)}.

Apart from our analysis on empirical NTK shown in Figure~\ref{figure:procgen_ntk_stats},
we provide additional NTK analysis by including TRAC in comparison and also showing the summation of the diagonal values of empirical NTK matrix.
The results are shown in Figure~\ref{figure:procgen_ntk_stats_appendix}.
C-CHAIN not only suppresses the off-diagonal values to decorrelate gradients, but also suppresses the diagonal values to prevent the increased scale of gradient, which is also deemed to be a pathology of neural network.
Moreover, we found TRAC lies between C-CHAIN and Vanilla. This indicates that the NTK statistics used in our work also explain the efficacy of TRAC.

\clearpage
\section{Complete Learning Curves}
\label{app:complete_learning_curves}

\begin{figure}[!h]
\begin{center}
\includegraphics[width=0.8\textwidth]{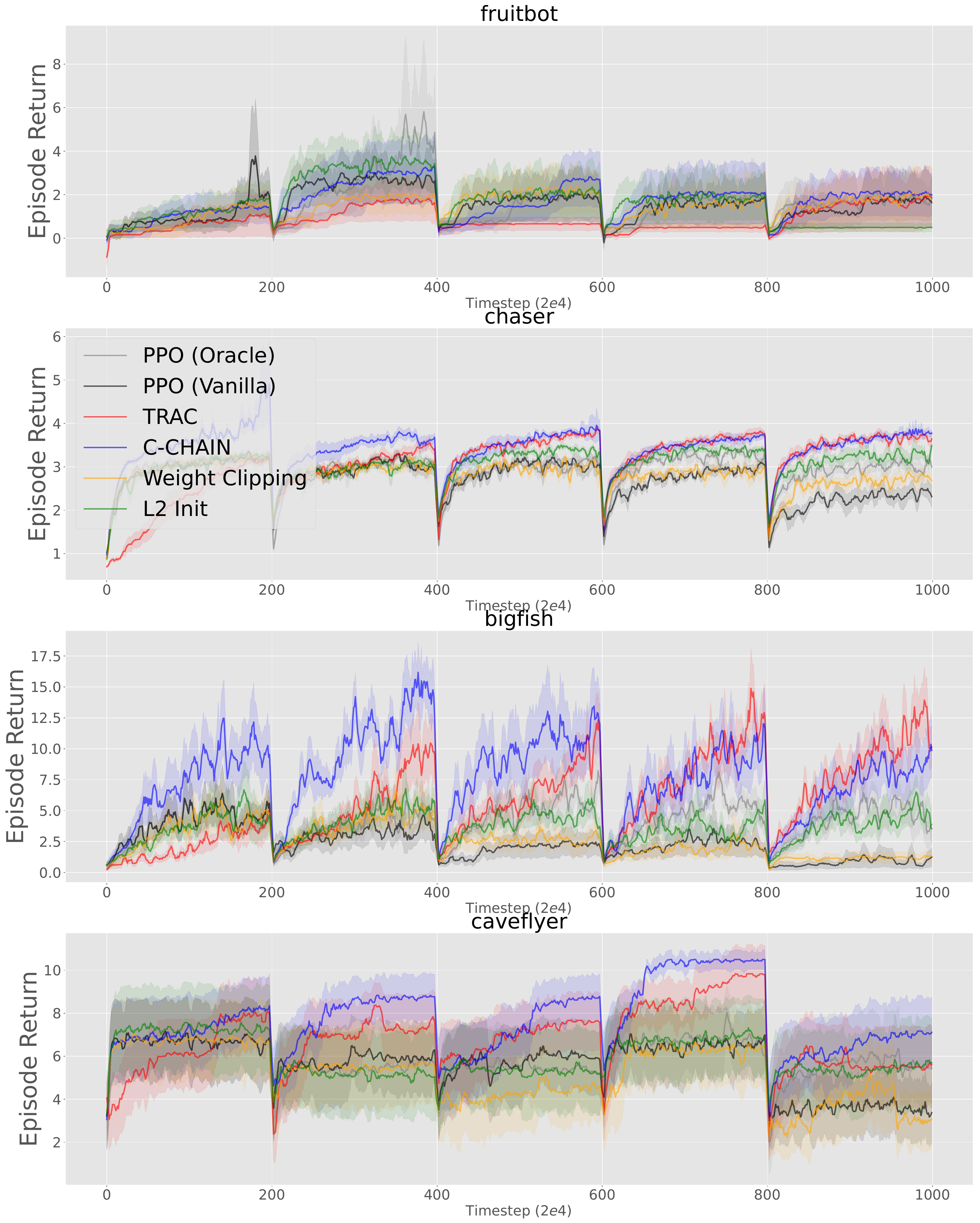}
\end{center}
\vspace{-0.3cm}
\caption{\textbf{Learning curves of different methods in four (of sixteen) continual ProcGen environments}: fruitbot, chaser, bigfish, caveflyer. The curves and the shades are means and standard errors over six seeds.}
\vspace{-0.5cm}
\label{figure:procgen_eval_n5_to_8}
\end{figure}

\begin{figure}
\begin{center}
\includegraphics[width=0.8\textwidth]{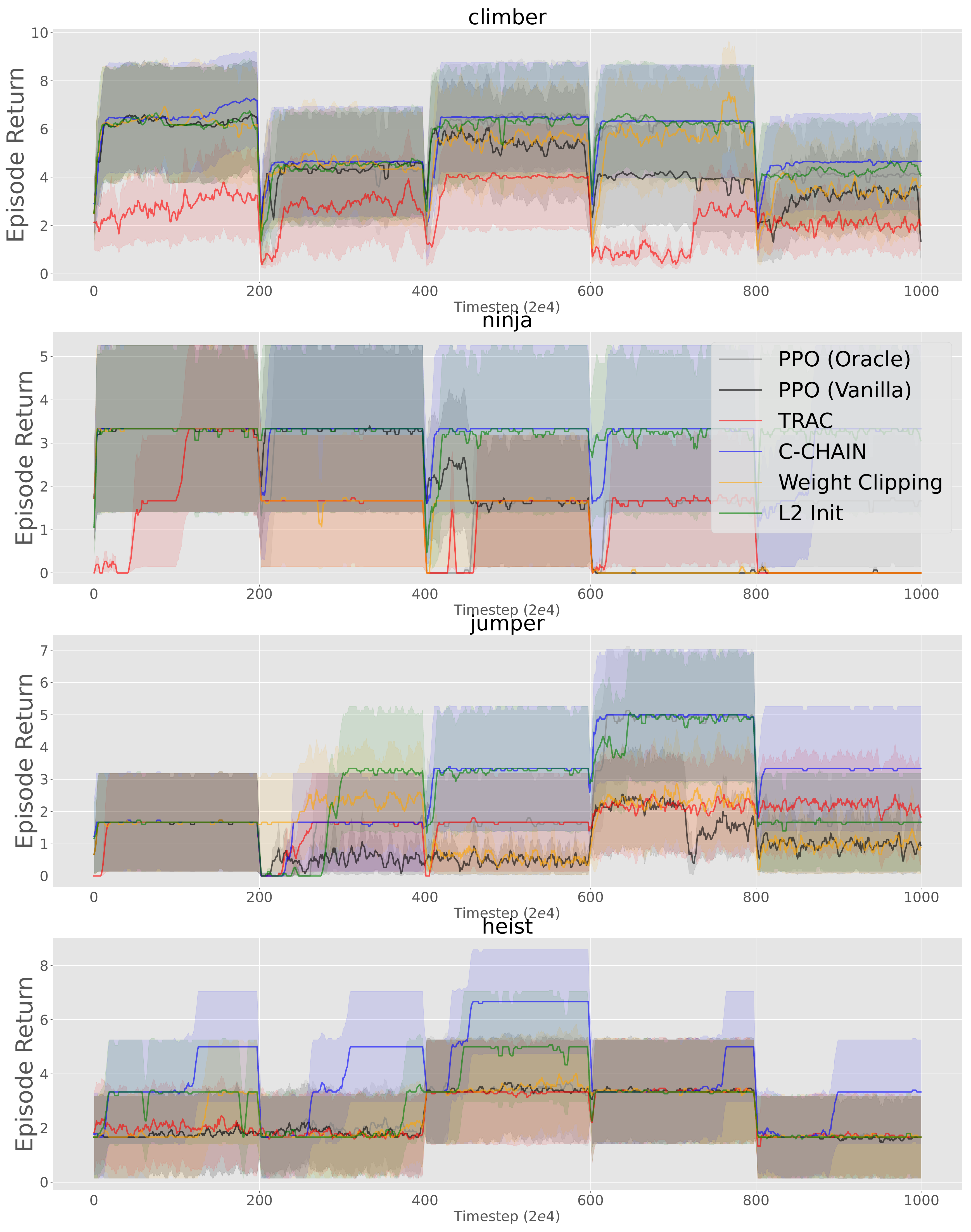}
\end{center}
\vspace{-0.3cm}
\caption{\textbf{Learning curves of different methods in four (of sixteen) continual ProcGen environments}: climber, ninja, jumper, heist. The curves and the shades are means and standard errors over six seeds.}
\vspace{-0.5cm}
\label{figure:procgen_eval_n9_to_12}
\end{figure}

\begin{figure}
\begin{center}
\includegraphics[width=0.8\textwidth]{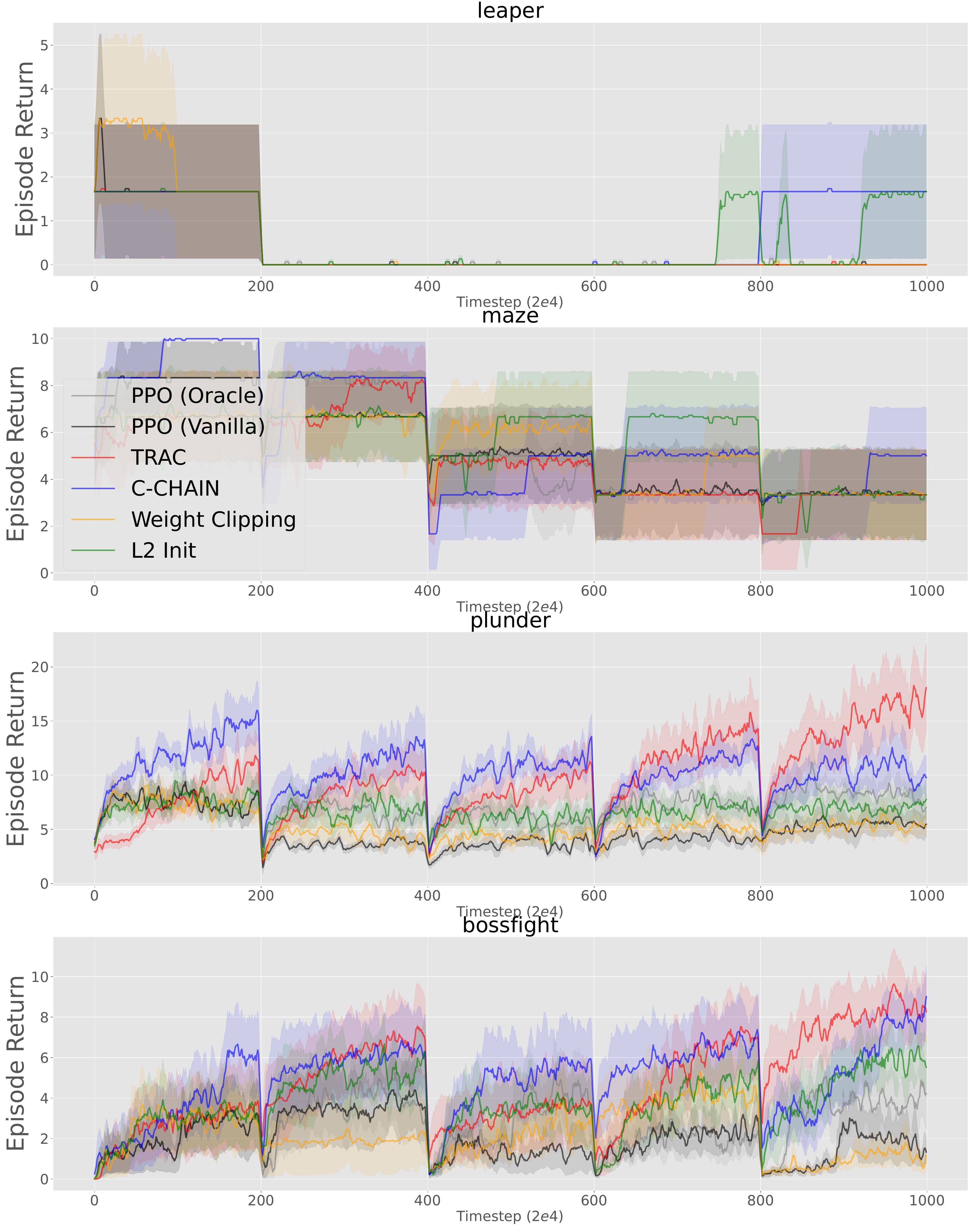}
\end{center}
\vspace{-0.3cm}
\caption{\textbf{Learning curves of different methods in four (of sixteen) continual ProcGen environments}: leaper, maze, plunder, bossfight. The curves and the shades are means and standard errors over six seeds.}
\vspace{-0.5cm}
\label{figure:procgen_eval_n13_to_16}
\end{figure}

The learning curves for the missing twelve contiunal ProcGen environments are provided in Figure~\ref{figure:procgen_eval_n5_to_8}, Figure~\ref{figure:procgen_eval_n9_to_12} and Figure~\ref{figure:procgen_eval_n13_to_16}.

\end{document}